%% file: paper.tex
\documentclass[letterpaper]{article} 
\usepackage{aaai24}  
\usepackage{times}  
\usepackage{helvet}  
\usepackage{courier}  
\usepackage[hyphens]{url}  
\usepackage{graphicx} 
\usepackage{natbib}  
\usepackage{caption} 
\usepackage{algorithm}

\usepackage{newfloat}
\usepackage{listings}
\DeclareCaptionStyle{ruled}{labelfont=normalfont,labelsep=colon,strut=off} 
\floatstyle{ruled}
\newfloat{listing}{tb}{lst}{}
\floatname{listing}{Listing}
\usepackage{amsfonts}       
\usepackage{amsmath}
\usepackage{amsthm}
\usepackage{caption}
\usepackage{graphicx}
\usepackage{algpseudocode}
\usepackage{import}
\usepackage{booktabs}
\usepackage{longtable}
\usepackage{array}
\usepackage{multirow}
\usepackage{placeins}

\newtheorem{definition}{Definition}[section]
\newtheorem{question}{Research Question}[section]

\title{Faithful Model Explanations through\\
Energy-Constrained Conformal Counterfactuals\footnote{This is a pre-print.}}
\author{
    Patrick Altmeyer\textsuperscript{\rm 1},
    Mojtaba Farmanbar\textsuperscript{\rm 2},
    Arie van Deursen\textsuperscript{\rm 1},
    Cynthia C. S. Liem\textsuperscript{\rm 1}
}
\affiliations{
    \textsuperscript{\rm 1}Delft University of Technology,\\
    Faculty of Electrical Engineering, Mathematics and Computer Science\\
    \textsuperscript{\rm 2}ING Bank\\
    p.altmeyer@tudelft.nl
}

\begin{document}

\maketitle

\begin{abstract}
  Counterfactual explanations offer an intuitive and straightforward way to explain black-box models and offer algorithmic recourse to individuals. To address the need for plausible explanations, existing work has primarily relied on surrogate models to learn how the input data is distributed. This effectively reallocates the task of learning realistic explanations for the data from the model itself to the surrogate. Consequently, the generated explanations may seem plausible to humans but need not necessarily describe the behaviour of the black-box model faithfully. We formalise this notion of faithfulness through the introduction of a tailored evaluation metric and propose a novel algorithmic framework for generating \textbf{E}nergy-\textbf{C}onstrained \textbf{C}onformal \textbf{Co}unterfactuals that are only as plausible as the model permits. Through extensive empirical studies, we demonstrate that \textit{ECCCo} reconciles the need for faithfulness and plausibility. In particular, we show that for models with gradient access, it is possible to achieve state-of-the-art performance without the need for surrogate models. To do so, our framework relies solely on properties defining the black-box model itself by leveraging recent advances in energy-based modelling and conformal prediction. To our knowledge, this is the first venture in this direction for generating faithful counterfactual explanations. Thus, we anticipate that \textit{ECCCo} can serve as a baseline for future research. We believe that our work opens avenues for researchers and practitioners seeking tools to better distinguish trustworthy from unreliable models.
\end{abstract}

\import{partials/}{body.tex}

\FloatBarrier

\bibliography{aaai24,bib}

\onecolumn

\import{partials/}{appendix.tex}

\end{document}

%% file: partials/body.tex
\section{Introduction}\label{intro}

Counterfactual explanations provide a powerful, flexible and intuitive way to not only explain black-box models but also offer the possibility of algorithmic recourse to affected individuals. Instead of opening the black box, counterfactual explanations work under the premise of strategically perturbing model inputs to understand model behaviour~\citep{wachter2017counterfactual}. Intuitively speaking, we generate explanations in this context by asking what-if questions of the following nature: `Our credit risk model currently predicts that this individual is not credit-worthy. What if they reduced their monthly expenditures by 10\%?'

This is typically implemented by defining a target outcome $\mathbf{y}^+ \in \mathcal{Y}$ for some individual $\mathbf{x} \in \mathcal{X}=\mathbb{R}^D$ described by $D$ attributes, for which the model $M_{\theta}:\mathcal{X}\mapsto\mathcal{Y}$ initially predicts a different outcome: $M_{\theta}(\mathbf{x})\ne \mathbf{y}^+$. Counterfactuals are then searched by minimizing a loss function that compares the predicted model output to the target outcome: $\text{yloss}(M_{\theta}(\mathbf{x}),\mathbf{y}^+)$. Since counterfactual explanations work directly with the black-box model, valid counterfactuals always have full local fidelity by construction where fidelity is defined as the degree to which explanations approximate the predictions of a black-box model~\citep{molnar2022interpretable}. 

In situations where full fidelity is a requirement, counterfactual explanations offer a more appropriate solution to Explainable Artificial Intelligence (XAI) than other popular approaches like LIME~\citep{ribeiro2016why} and SHAP~\citep{lundberg2017unified}, which involve local surrogate models. But even full fidelity is not a sufficient condition for ensuring that an explanation \textit{faithfully} describes the behaviour of a model. That is because multiple distinct explanations can lead to the same model prediction, especially when dealing with heavily parameterized models like deep neural networks, which are underspecified by the data~\citep{wilson2020case}. In the context of counterfactuals, the idea that no two explanations are the same arises almost naturally. A key focus in the literature has therefore been to identify those explanations that are most appropriate based on a myriad of desiderata such as closeness~\citep{wachter2017counterfactual}, sparsity~\citep{schut2021generating}, actionability~\citep{ustun2019actionable} and plausibility~\citep{joshi2019realistic}. 

In this work, we draw closer attention to modelling faithfulness rather than fidelity as a desideratum for counterfactuals. We define faithfulness as the degree to which counterfactuals are consistent with what the model has learned about the data. Our key contributions are as follows: first, we show that fidelity is an insufficient evaluation metric for counterfactuals (Section~\ref{fidelity}) and propose a definition of faithfulness that gives rise to more suitable metrics (Section~\ref{faithfulness}). Next, we introduce a \textit{ECCCo}: a novel algorithmic approach aimed at generating energy-constrained conformal counterfactuals that faithfully explain model behaviour in Section~\ref{meth}. Finally, we provide extensive empirical evidence demonstrating that \textit{ECCCo} faithfully explains model behaviour and attains plausibility only when appropriate (Section~\ref{emp}).

To our knowledge, this is the first venture in this direction for generating faithful counterfactuals. Thus, we anticipate that \textit{ECCCo} can serve as a baseline for future research. We believe that our work opens avenues for researchers and practitioners seeking tools to better distinguish trustworthy from unreliable models.

\section{Background}\label{background}

While counterfactual explanations (CE) can also be generated for arbitrary regression models~\citep{spooner2021counterfactual}, existing work has primarily focused on classification problems. Let $\mathcal{Y}=(0,1)^K$ denote the one-hot-encoded output domain with $K$ classes. Then most counterfactual generators rely on gradient descent to optimize different flavours of the following counterfactual search objective:

\begin{equation} \label{eq:general}
\begin{aligned}
\min_{\mathbf{Z}^\prime \in \mathcal{Z}^L} \left\{  {\text{yloss}(M_{\theta}(f(\mathbf{Z}^\prime)),\mathbf{y}^+)}+ \lambda {\text{cost}(f(\mathbf{Z}^\prime)) }  \right\} 
\end{aligned} 
\end{equation}

Here $\text{yloss}(\cdot)$ denotes the primary loss function, $f(\cdot)$ is a function that maps from the counterfactual state space to the feature space and $\text{cost}(\cdot)$ is either a single penalty or a collection of penalties that are used to impose constraints through regularization. Equation~\ref{eq:general} restates the baseline approach to gradient-based counterfactual search proposed by~\citet{wachter2017counterfactual} in general form as introduced by~\citet{altmeyer2023endogenous}. To explicitly account for the multiplicity of explanations, $\mathbf{Z}^\prime=\{ \mathbf{z}_l\}_L$ denotes an $L$-dimensional array of counterfactual states. 

The baseline approach, which we will simply refer to as \textit{Wachter}, searches a single counterfactual directly in the feature space and penalizes its distance to the original factual. In this case, $f(\cdot)$ is simply the identity function and $\mathcal{Z}$ corresponds to the feature space itself. Many derivative works of~\citet{wachter2017counterfactual} have proposed new flavours of Equation~\ref{eq:general}, each of them designed to address specific \textit{desiderata} that counterfactuals ought to meet in order to properly serve both AI practitioners and individuals affected by algorithmic decision-making systems. The list of desiderata includes but is not limited to the following: sparsity, closeness~\citep{wachter2017counterfactual}, actionability~\citep{ustun2019actionable}, diversity~\citep{mothilal2020explaining}, plausibility~\citep{joshi2019realistic,poyiadzi2020face,schut2021generating}, robustness~\citep{upadhyay2021robust,pawelczyk2022probabilistically,altmeyer2023endogenous} and causality~\citep{karimi2021algorithmic}. Different counterfactual generators addressing these needs have been extensively surveyed and evaluated in various studies~\citep{verma2020counterfactual,karimi2020survey,pawelczyk2021carla,artelt2021evaluating,guidotti2022counterfactual}. 

The notion of plausibility is central to all of the desiderata. For example, \citet{artelt2021evaluating} find that plausibility typically also leads to improved robustness. Similarly, plausibility has also been connected to causality in the sense that plausible counterfactuals respect causal relationships~\citep{mahajan2019preserving}. Consequently, the plausibility of counterfactuals has been among the primary concerns for researchers. Achieving plausibility is equivalent to ensuring that the generated counterfactuals comply with the true and unobserved data-generating process (DGP). We define plausibility formally in this work as follows:

\begin{definition}[Plausible Counterfactuals]
  \label{def:plausible}
  Let $\mathcal{X}|\mathbf{y}^+= p(\mathbf{x}|\mathbf{y}^+)$ denote the true conditional distribution of samples in the target class $\mathbf{y}^+$. Then for $\mathbf{x}^{\prime}$ to be considered a plausible counterfactual, we need: $\mathbf{x}^{\prime} \sim \mathcal{X}|\mathbf{y}^+$.
\end{definition}

To generate plausible counterfactuals, we first need to quantify the conditional distribution of samples in the target class ($\mathcal{X}|\mathbf{y}^+$). We can then ensure that we generate counterfactuals that comply with that distribution.

One straightforward way to do this is to use surrogate models for the task. \citet{joshi2019realistic}, for example, suggest that instead of searching counterfactuals in the feature space $\mathcal{X}$, we can traverse a latent embedding $\mathcal{Z}$ (Equation~\ref{eq:general}) that implicitly codifies the DGP. To learn the latent embedding, they propose using a generative model such as a Variational Autoencoder (VAE). Provided the surrogate model is well-specified, their proposed approach \textit{REVISE} can yield plausible explanations. Others have proposed similar approaches: \citet{dombrowski2021diffeomorphic} traverse the base space of a normalizing flow to solve Equation~\ref{eq:general}; \citet{poyiadzi2020face} use density estimators ($\hat{p}: \mathcal{X} \mapsto [0,1]$) to constrain the counterfactuals to dense regions in the feature space; finally, \citet{karimi2021algorithmic} assume knowledge about the causal graph that generates the data.

A competing approach towards plausibility that is also closely related to this work instead relies on the black-box model itself.~\citet{schut2021generating} show that to meet the plausibility objective we need not explicitly model the input distribution. Pointing to the undesirable engineering overhead induced by surrogate models, they propose to rely on the implicit minimization of predictive uncertainty instead. Their proposed methodology, which we will refer to as \textit{Schut}, solves Equation~\ref{eq:general} by greedily applying Jacobian-Based Saliency Map Attacks (JSMA) in the feature space with cross-entropy loss and no penalty at all. The authors demonstrate theoretically and empirically that their approach yields counterfactuals for which the model $M_{\theta}$ predicts the target label $\mathbf{y}^+$ with high confidence. Provided the model is well-specified, these counterfactuals are plausible. This idea hinges on the assumption that the black-box model provides well-calibrated predictive uncertainty estimates.

\section{Why Fidelity is not Enough: A Motivational Example}\label{fidelity}

As discussed in the introduction, any valid counterfactual also has full fidelity by construction: solutions to Equation~\ref{eq:general} are considered valid as soon as the label predicted by the model matches the target class. So while fidelity always applies, counterfactuals that address the various desiderata introduced above can look vastly different from each other. 

\begin{figure}
  \centering
  \includegraphics[width=0.8\linewidth]{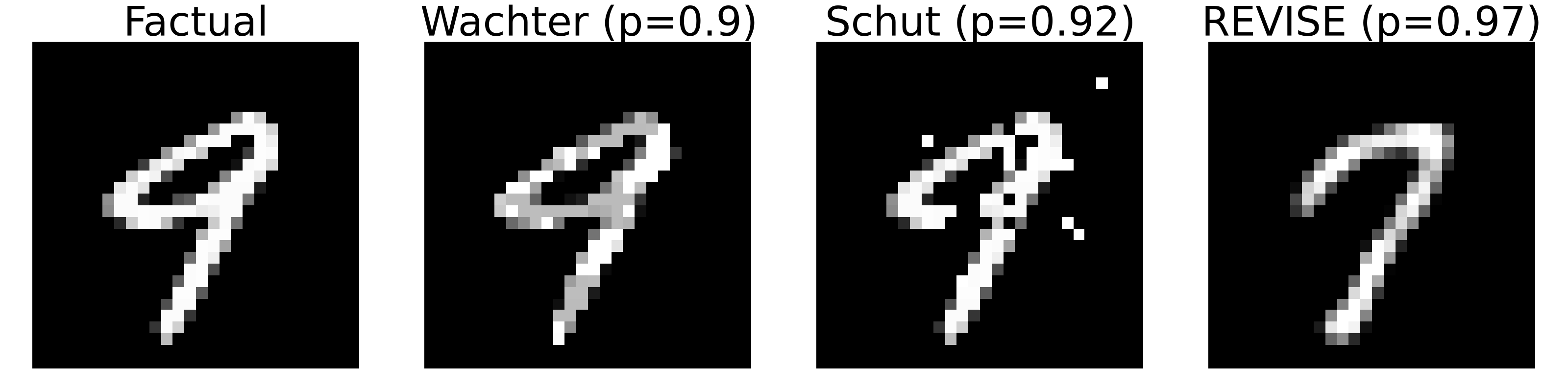}
  \caption{Counterfactuals for turning a 9 (nine) into a 7 (seven): original image (left), then the counterfactuals generated using \textit{Wachter}, \textit{Schut} and \textit{REVISE}.}\label{fig:motiv}
\end{figure}

To demonstrate this with an example, we have trained a simple image classifier $M_{\theta}$ on the well-known \textit{MNIST} dataset~\citep{lecun1998mnist}: a Multi-Layer Perceptron (\textit{MLP}) with test set accuracy $> 0.9$. No measures have been taken to improve the model's adversarial robustness or its capacity for predictive uncertainty quantification. The far left panel of Figure~\ref{fig:motiv} shows a random sample drawn from the dataset. The underlying classifier correctly predicts the label `nine' for this image. For the given factual image and model, we have used \textit{Wachter}, \textit{Schut} and \textit{REVISE} to generate one counterfactual each in the target class `seven'. The perturbed images are shown next to the factual image from left to right in Figure~\ref{fig:motiv}. Captions on top of the images indicate the generator along with the predicted probability that the image belongs to the target class. In all cases, that probability is very high, while the counterfactuals look very different.

Since \textit{Wachter} is only concerned with closeness, the generated counterfactual is almost indistinguishable from the factual. \textit{Schut} expects a well-calibrated model that can generate predictive uncertainty estimates. Since this is not the case, the generated counterfactual looks like an adversarial example. Finally, the counterfactual generated by \textit{REVISE} looks much more plausible than the other two. But is it also more faithful to the behaviour of our \textit{MNIST} classifier? That is much less clear because the surrogate used by \textit{REVISE} introduces friction: explanations no longer depend exclusively on the black-box model itself. 

So which of the counterfactuals most faithfully explains the behaviour of our image classifier? Fidelity cannot help us to make that judgement, because all of these counterfactuals have full fidelity. Thus, fidelity is an insufficient evaluation metric to assess the faithfulness of CE. 

\section{Faithful first, Plausible second}\label{faithfulness}

Considering the limitations of fidelity as demonstrated in the previous section, analogous to Definition~\ref{def:plausible}, we introduce a new notion of faithfulness in the context of CE:

\begin{definition}[Faithful Counterfactuals]
  \label{def:faithful}
  Let $\mathcal{X}_{\theta}|\mathbf{y}^+ = p_{\theta}(\mathbf{x}|\mathbf{y}^+)$ denote the conditional distribution of $\mathbf{x}$ in the target class $\mathbf{y}^+$, where $\theta$ denotes the parameters of model $M_{\theta}$. Then for $\mathbf{x}^{\prime}$ to be considered a faithful counterfactual, we need: $\mathbf{x}^{\prime} \sim \mathcal{X}_{\theta}|\mathbf{y}^+$.
\end{definition}

In doing this, we merge in and nuance the concept of plausibility (Definition~\ref{def:plausible}) where the notion of `consistent with the data' becomes `consistent with what the model has learned about the data'.

\subsection{Quantifying the Model's Generative Property}

To assess counterfactuals with respect to Definition~\ref{def:faithful}, we need a way to quantify the posterior conditional distribution $p_{\theta}(\mathbf{x}|\mathbf{y}^+)$. To this end, we draw on ideas from energy-based modelling (EBM), a subdomain of machine learning that is concerned with generative or hybrid modelling~\citep{grathwohl2020your,du2019implicit}. In particular, note that if we fix $\mathbf{y}$ to our target value $\mathbf{y}^+$, we can conditionally draw from $p_{\theta}(\mathbf{x}|\mathbf{y}^+)$ by randomly initializing $\mathbf{x}_0$ and then using Stochastic Gradient Langevin Dynamics (SGLD) as follows, 

\begin{equation}\label{eq:sgld}
  \begin{aligned}
    \mathbf{x}_{j+1} &\leftarrow \mathbf{x}_j - \frac{\epsilon_j^2}{2} \mathcal{E}_{\theta}(\mathbf{x}_j|\mathbf{y}^+) + \epsilon_j \mathbf{r}_j, && j=1,...,J
  \end{aligned}
\end{equation}

where $\mathbf{r}_j \sim \mathcal{N}(\mathbf{0},\mathbf{I})$ is the stochastic term and the step-size $\epsilon_j$ is typically polynomially decayed~\citep{welling2011bayesian}. The term $\mathcal{E}_{\theta}(\mathbf{x}_j|\mathbf{y}^+)$ denotes the model energy conditioned on the target class label $\mathbf{y}^+$ which we specify as the negative logit corresponding to $\mathbf{y}^{+}$. To allow for faster sampling, we follow the common practice of choosing the step-size $\epsilon_j$ and the standard deviation of $\mathbf{r}_j$ separately. While $\mathbf{x}_J$ is only guaranteed to distribute as $p_{\theta}(\mathbf{x}|\mathbf{y}^{+})$ if $\epsilon \rightarrow 0$ and $J \rightarrow \infty$, the bias introduced for a small finite $\epsilon$ is negligible in practice \citep{murphy2023probabilistic}. 

Generating multiple samples using SGLD thus yields an empirical distribution $\widehat{\mathbf{X}}_{\theta,\mathbf{y}^+}$ that approximates what the model has learned about the input data. While in the context of EBM, this is usually done during training, we propose to repurpose this approach during inference in order to evaluate the faithfulness of model explanations. The appendix provides additional implementation details for any tasks related to energy-based modelling. 

\subsection{Quantifying the Model's Predictive Uncertainty}

Faithful counterfactuals can be expected to also be plausible if the learned conditional distribution $\mathcal{X}_{\theta}|\mathbf{y}^+$ (Defintion~\ref{def:faithful}) is close to the true conditional distribution $\mathcal{X}|\mathbf{y}^+$ (Definition~\ref{def:plausible}). We can further improve the plausibility of counterfactuals without the need for surrogate models that may interfere with faithfulness by minimizing predictive uncertainty~\citep{schut2021generating}.
Unfortunately, this idea relies on the assumption that the model itself provides predictive uncertainty estimates, which may be too restrictive in practice. 

To relax this assumption, we use conformal prediction (CP), an approach to predictive uncertainty quantification that has recently gained popularity~\citep{angelopoulos2021gentle,manokhin2022awesome}. Crucially for our intended application, CP is model-agnostic and can be applied during inference without placing any restrictions on model training. It works under the premise of turning heuristic notions of uncertainty into rigorous estimates by repeatedly sifting through the training data or a dedicated calibration dataset. Calibration data is used to compute so-called nonconformity scores: $\mathcal{S}=\{s(\mathbf{x}_i,\mathbf{y}_i)\}_{i \in \mathcal{D}_{\text{cal}}}$ where $s: (\mathcal{X},\mathcal{Y}) \mapsto \mathbb{R}$ is referred to as \textit{score function} (see appendix for details).

Conformal classifiers produce prediction sets for individual inputs that include all output labels that can be reasonably attributed to the input. These sets are formed as follows,

\begin{equation}\label{eq:scp}
  \begin{aligned}
    C_{\theta}(\mathbf{x}_i;\alpha)=\{\mathbf{y}: s(\mathbf{x}_i,\mathbf{y}) \le \hat{q}\}
  \end{aligned}
\end{equation}

where $\hat{q}$ denotes the $(1-\alpha)$-quantile of $\mathcal{S}$ and $\alpha$ is a predetermined error rate. These sets tend to be larger for inputs that do not conform with the training data and are characterized by high predictive uncertainty. To leverage this notion of predictive uncertainty in the context of gradient-based counterfactual search, we use a smooth set size penalty introduced by~\citet{stutz2022learning}:

\begin{equation}\label{eq:setsize}
  \begin{aligned}
    \Omega(C_{\theta}(\mathbf{x};\alpha))&=\max \left(0, \sum_{\mathbf{y}\in\mathcal{Y}}C_{\theta,\mathbf{y}}(\mathbf{x}_i;\alpha) - \kappa \right)
  \end{aligned}
\end{equation}

Here, $\kappa \in \{0,1\}$ is a hyper-parameter and $C_{\theta,\mathbf{y}}(\mathbf{x}_i;\alpha)$ can be interpreted as the probability of label $\mathbf{y}$ being included in the prediction set (see appendix for details). In order to compute this penalty for any black-box model, we merely need to perform a single calibration pass through a holdout set $\mathcal{D}_{\text{cal}}$. Arguably, data is typically abundant and in most applications, practitioners tend to hold out a test data set anyway. Consequently, CP removes the restriction on the family of predictive models, at the small cost of reserving a subset of the available data for calibration. This particular case of conformal prediction is referred to as \textit{split conformal prediction} (SCP) as it involves splitting the training data into a proper training dataset and a calibration dataset.

\subsection{Evaluating Plausibility and Faithfulness}

The parallels between our definitions of plausibility and faithfulness imply that we can also use similar evaluation metrics in both cases. Since existing work has focused heavily on plausibility, it offers a useful starting point. In particular,~\citet{guidotti2022counterfactual} have proposed an implausibility metric that measures the distance of the counterfactual from its nearest neighbour in the target class. As this distance is reduced, counterfactuals get more plausible under the assumption that the nearest neighbour itself is plausible in the sense of Definition~\ref{def:plausible}. In this work, we use the following adapted implausibility metric,

\begin{equation}\label{eq:impl}
  \begin{aligned}
    \text{impl}(\mathbf{x}^{\prime},\mathbf{X}_{\mathbf{y}^+}) = \frac{1}{\lvert\mathbf{X}_{\mathbf{y}^+}\rvert} \sum_{\mathbf{x} \in \mathbf{X}_{\mathbf{y}^+}} \text{dist}(\mathbf{x}^{\prime},\mathbf{x})
  \end{aligned}
\end{equation}

where $\mathbf{x}^{\prime}$ denotes the counterfactual and $\mathbf{X}_{\mathbf{y}^+}$ is a subsample of the training data in the target class $\mathbf{y}^+$. By averaging over multiple samples in this manner, we avoid the risk that the nearest neighbour of $\mathbf{x}^{\prime}$ itself is not plausible according to Definition~\ref{def:plausible} (e.g. an outlier).

Equation~\ref{eq:impl} gives rise to a similar evaluation metric for unfaithfulness. We swap out the subsample of observed individuals in the target class for the set of samples generated through SGLD ($\widehat{\mathbf{X}}_{\theta,\mathbf{y}^+}$):

\begin{equation}\label{eq:faith}
  \begin{aligned}
    \text{unfaith}(\mathbf{x}^{\prime},\widehat{\mathbf{X}}_{\theta,\mathbf{y}^+}) = \frac{1}{\lvert \widehat{\mathbf{X}}_{\theta,\mathbf{y}^+} \rvert} \sum_{\mathbf{x} \in \widehat{\mathbf{X}}_{\theta,\mathbf{y}^+}} \text{dist}(\mathbf{x}^{\prime},\mathbf{x})
  \end{aligned}
\end{equation}

Our default choice for the $\text{dist}(\cdot)$ function in both cases is the Euclidean Norm. Depending on the type of input data other choices may be more adequate (see Section~\ref{emp:setup}). 

\section{Energy-Constrained Conformal Counterfactuals}\label{meth}

Given our proposed notion of faithfulness, we now describe \textit{ECCCo}, our proposed framework for generating Energy-Constrained Conformal Counterfactuals. It is based on the premise that counterfactuals should first and foremost be faithful. Plausibility, as a secondary concern, is then still attainable to the degree that the black-box model itself has learned plausible explanations for the underlying data. 

We begin by substituting the loss function in Equation~\ref{eq:general},

\begin{equation} \label{eq:eccco-start}
  \begin{aligned}
  \min_{\mathbf{Z}^\prime \in \mathcal{Z}^L} \{  {L_{\text{JEM}}(f(\mathbf{Z}^\prime);M_{\theta},\mathbf{y}^+)}+ \lambda {\text{cost}(f(\mathbf{Z}^\prime)) } \} 
  \end{aligned} 
\end{equation}

where $L_{\text{JEM}}(f(\mathbf{Z}^\prime);M_{\theta},\mathbf{y}^+)$ is a hybrid loss function used in joint-energy modelling evaluated at a given counterfactual state for a given model and target outcome:

\begin{equation}
  \begin{aligned}
    L_{\text{JEM}}(f(\mathbf{Z}^\prime); \cdot) = L_{\text{clf}}(f(\mathbf{Z}^\prime); \cdot) + L_{\text{gen}}(f(\mathbf{Z}^\prime); \cdot)
  \end{aligned}
\end{equation}

The first term, $L_{\text{clf}}$, is any standard classification loss function such as cross-entropy loss. The second term, $L_{\text{gen}}$, is used to measure loss with respect to the generative task\footnote{In practice, regularization loss is typically also added. We follow this convention but have omitted the term here for simplicity.}. In the context of joint-energy training, $L_{\text{gen}}$ induces changes in model parameters $\theta$ that decrease the energy of observed samples and increase the energy of samples generated through SGLD~\citep{du2019implicit}. 

The key observation in our context is that we can rely solely on decreasing the energy of the counterfactual itself. This is sufficient to capture the generative property of the underlying model since it is implicitly captured by its parameters $\theta$. Importantly, this means that we do not need to generate conditional samples through SGLD during our counterfactual search at all (see appendix for details).

This observation leads to the following simple objective function for \textit{ECCCo}:

\begin{equation} \label{eq:eccco}
  \begin{aligned}
  & \min_{\mathbf{Z}^\prime \in \mathcal{Z}^L} \{  {L_{\text{clf}}(f(\mathbf{Z}^\prime);M_{\theta},\mathbf{y}^+)}+ \lambda_1 {\text{cost}(f(\mathbf{Z}^\prime)) } \\
  &+ \lambda_2 \mathcal{E}_{\theta}(f(\mathbf{Z}^\prime)|\mathbf{y}^+) + \lambda_3 \Omega(C_{\theta}(f(\mathbf{Z}^\prime);\alpha)) \} 
  \end{aligned} 
\end{equation}

The first penalty term involving $\lambda_1$ induces closeness like in~\citet{wachter2017counterfactual}. The second penalty term involving $\lambda_2$ induces faithfulness by constraining the energy of the generated counterfactual. The third and final penalty term involving $\lambda_3$ ensures that the generated counterfactual is associated with low predictive uncertainty. To tune these hyperparameters we have relied on grid search.

Concerning feature autoencoding ($f: \mathcal{Z} \mapsto \mathcal{X}$), \textit{ECCCo} does not rely on latent space search to achieve its primary objective of faithfulness. By default, we choose $f(\cdot)$ to be the identity function as in \textit{Wachter}. This is generally also enough to achieve plausibility, provided the model has learned plausible explanations for the data. In some cases, plausibility can be improved further by mapping counterfactuals to a lower-dimensional latent space. In the following, we refer to this approach as \textit{ECCCo+}: that is, \textit{ECCCo} plus dimensionality reduction.

\begin{figure*}
  \centering
  \includegraphics[width=0.75\linewidth]{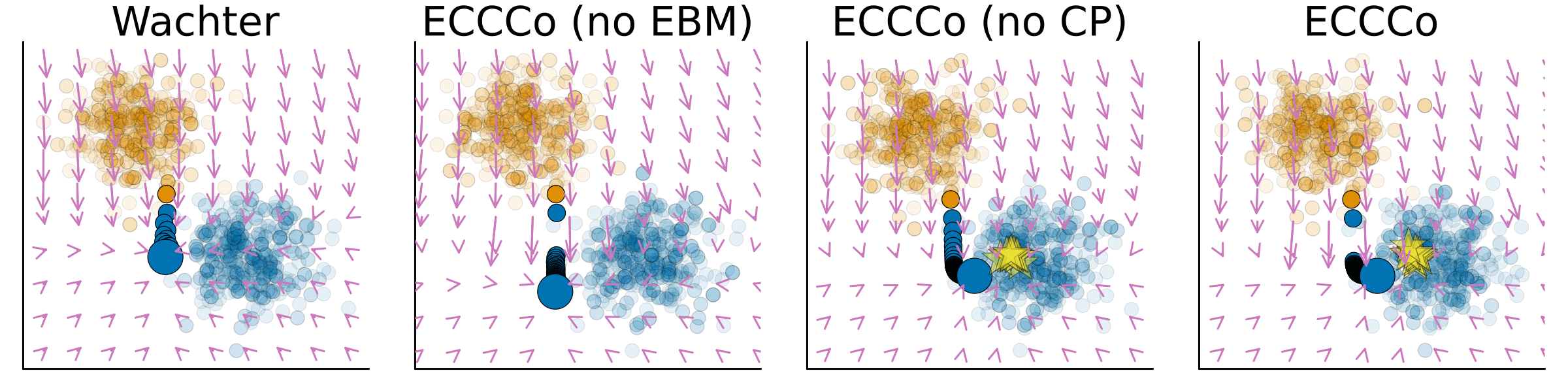}
  \caption{Gradient fields and counterfactual paths for different generators. The objective is to generate a counterfactual in the blue class for a sample from the orange class. Bright yellow stars indicate conditional samples generated through SGLD. The underlying classifier is a Joint Energy Model.}\label{fig:poc}
\end{figure*}  

Figure~\ref{fig:poc} illustrates how the different components in Equation~\ref{eq:eccco} affect the counterfactual search for a synthetic dataset. The underlying classifier is a Joint Energy Model (\textit{JEM}) that was trained to predict the output class (blue or orange) and generate class-conditional samples~\citep{grathwohl2020your}. We have used four different generator flavours to produce a counterfactual in the blue class for a sample from the orange class: \textit{Wachter}, which only uses the first penalty ($\lambda_2=\lambda_3=0$); \textit{ECCCo (no EBM)}, which does not constrain energy ($\lambda_2=0$); \textit{ECCCo (no CP)}, which involves no set size penalty ($\lambda_3=0$); and, finally, \textit{ECCCo}, which involves all penalties defined in Equation~\ref{eq:eccco}. Arrows indicate (negative) gradients with respect to the objective function at different points in the feature space. 

While \textit{Wachter} generates a valid counterfactual, it ends up close to the original starting point consistent with its objective. \textit{ECCCo (no EBM)} avoids regions of high predictive uncertainty near the decision boundary, but the outcome is still not plausible. The counterfactual produced by \textit{ECCCo (no CP)} is energy-constrained. Since the \textit{JEM} has learned the conditional input distribution reasonably well in this case, the counterfactual is both faithful and plausible. Finally, the outcome for \textit{ECCCo} looks similar, but the additional smooth set size penalty leads to somewhat faster convergence. 

\section{Empirical Analysis}\label{emp}

Our goal in this section is to shed light on the following research questions:

\begin{question}[Faithfulness]\label{rq:faithfulness}
  To what extent are counterfactuals generated by \textit{ECCCo} more faithful than those produced by state-of-the-art generators?
\end{question}

\begin{question}[Balancing Desiderata]\label{rq:plausibility}
  Compared to state-of-the-art generators, how does \textit{ECCCo} balance the two key objectives of faithfulness and plausibility?
\end{question}

The second question is motivated by the intuition that faithfulness and plausibility should coincide for models that have learned plausible explanations of the data.

\subsection{Experimental Setup}\label{emp:setup}

To assess and benchmark the performance of our proposed generator against the state of the art, we generate multiple counterfactuals for different models and datasets. In particular, we compare \textit{ECCCo} and its variants to the following counterfactual generators that were introduced above: firstly, \textit{Schut}, which works under the premise of minimizing predictive uncertainty; secondly, \textit{REVISE}, which is state-of-the-art (SOTA) with respect to plausibility; and, finally, \textit{Wachter}, which serves as our baseline. In the case of \textit{ECCCo+}, we use principal component analysis (PCA) for dimensionality reduction: the latent space $\mathcal{Z}$ is spanned by the first $n_z$ principal components where we choose $n_z$ to be equal to the latent dimension of the VAE used by \textit{REVISE}.

For the predictive modelling tasks, we use multi-layer perceptrons (\textit{MLP}), deep ensembles, joint energy models (\textit{JEM}) and convolutional neural networks (LeNet-5 \textit{CNN}~\citep{lecun1998gradient}). Both joint-energy modelling and ensembling have been associated with improved generative properties and adversarial robustness~\citep{grathwohl2020your,lakshminarayanan2016simple}, so we expect this to be positively correlated with the plausibility of \textit{ECCCo}. To account for stochasticity, we generate many counterfactuals for each target class, generator, model and dataset over multiple runs.

We perform benchmarks on eight datasets from different domains. From the credit and finance domain we include three tabular datasets: Give Me Some Credit (\textit{GMSC})~\citep{kaggle2011give}, \textit{German Credit}~\citep{hoffman1994german} and \textit{California Housing}~\citep{pace1997sparse}. All of these are commonly used in the related literature~\citep{karimi2020survey,altmeyer2023endogenous,pawelczyk2021carla}. Following related literature~\citep{schut2021generating,dhurandhar2018explanations} we also include two image datasets: \textit{MNIST}~\citep{lecun1998mnist} and \textit{Fashion MNIST}~\citep{xiao2017fashion}. 

Full details concerning model training as well as detailed descriptions and results for all datasets can be found in the appendix. In the following, we will focus on the most relevant results highlighted in Tables~\ref{tab:results-tabular} and~\ref{tab:results-vision}. The tables show sample averages along with standard deviations across multiple runs for our key evaluation metrics for the \textit{California Housing} and \textit{GMSC} datasets (Table~\ref{tab:results-tabular}) and the \textit{MNIST} dataset (Table~\ref{tab:results-vision}). For each metric, the best outcomes are highlighted in bold. Asterisks indicate that the given value is more than one (*) or two (**) standard deviations away from the baseline (Wachter). For the tabular datasets, we use the default Euclidian distance to measure unfaithfulness and implausibility as defined in Equations~\ref{eq:faith} and~\ref{eq:impl}, respectively. The third metric presented in Table~\ref{tab:results-tabular} quantifies the predictive uncertainty of the counterfactual as measured by Equation~\ref{eq:setsize}. For the vision datasets, we rely on measuring the structural dissimilarity between images for our unfaithfulness and implausibility metrics~\citep{wang2003multiscale}. 

\subsection{Faithfulness}

\begin{table*}
\centering
\resizebox{\linewidth}{!}{
\begin{tabular}[t]{llcccccc}
\toprule
\multicolumn{2}{c}{ } & \multicolumn{3}{c}{California Housing} & \multicolumn{3}{c}{GMSC} \\
\cmidrule(l{3pt}r{3pt}){3-5} \cmidrule(l{3pt}r{3pt}){6-8}
Model & Generator & Unfaithfulness ↓ & Implausibility ↓ & Uncertainty ↓ & Unfaithfulness ↓ & Implausibility ↓ & Uncertainty ↓\\
\midrule
    & ECCCo & \textbf{3.69 ± 0.08}** & 1.94 ± 0.13\hphantom{*}\hphantom{*} & \textbf{0.09 ± 0.01}** & 3.84 ± 0.07** & 2.13 ± 0.08\hphantom{*}\hphantom{*} & \textbf{0.23 ± 0.01}**\\

    & ECCCo+ & 3.88 ± 0.07** & 1.20 ± 0.09\hphantom{*}\hphantom{*} & 0.15 ± 0.02\hphantom{*}\hphantom{*} & \textbf{3.79 ± 0.05}** & 1.81 ± 0.05\hphantom{*}\hphantom{*} & 0.30 ± 0.01*\hphantom{*}\\

    & ECCCo (no CP) & 3.70 ± 0.08** & 1.94 ± 0.13\hphantom{*}\hphantom{*} & 0.10 ± 0.01** & 3.85 ± 0.07** & 2.13 ± 0.08\hphantom{*}\hphantom{*} & 0.23 ± 0.01**\\

    & ECCCo (no EBM) & 4.03 ± 0.07\hphantom{*}\hphantom{*} & 1.12 ± 0.12\hphantom{*}\hphantom{*} & 0.14 ± 0.01** & 4.08 ± 0.06\hphantom{*}\hphantom{*} & 0.97 ± 0.08\hphantom{*}\hphantom{*} & 0.31 ± 0.01*\hphantom{*}\\

    & REVISE & 3.96 ± 0.07*\hphantom{*} & \textbf{0.58 ± 0.03}** & 0.17 ± 0.03\hphantom{*}\hphantom{*} & 4.09 ± 0.07\hphantom{*}\hphantom{*} & \textbf{0.63 ± 0.02}** & 0.33 ± 0.06\hphantom{*}\hphantom{*}\\

    & Schut & 4.00 ± 0.06\hphantom{*}\hphantom{*} & 1.15 ± 0.12\hphantom{*}\hphantom{*} & 0.10 ± 0.01** & 4.04 ± 0.08\hphantom{*}\hphantom{*} & 1.21 ± 0.08\hphantom{*}\hphantom{*} & 0.30 ± 0.01*\hphantom{*}\\

\multirow{-7}{*}{\raggedright\arraybackslash MLP Ensemble} & Wachter & 4.04 ± 0.07\hphantom{*}\hphantom{*} & 1.13 ± 0.12\hphantom{*}\hphantom{*} & 0.16 ± 0.01\hphantom{*}\hphantom{*} & 4.10 ± 0.07\hphantom{*}\hphantom{*} & 0.95 ± 0.08\hphantom{*}\hphantom{*} & 0.32 ± 0.01\hphantom{*}\hphantom{*}\\
\cmidrule{1-8}
    & ECCCo & 1.40 ± 0.08** & 0.69 ± 0.05** & 0.11 ± 0.00** & 1.20 ± 0.06*\hphantom{*} & 0.78 ± 0.07** & 0.38 ± 0.01\hphantom{*}\hphantom{*}\\

    & ECCCo+ & \textbf{1.28 ± 0.08}** & 0.60 ± 0.04** & 0.11 ± 0.00** & \textbf{1.01 ± 0.07}** & 0.70 ± 0.07** & 0.37 ± 0.01\hphantom{*}\hphantom{*}\\

    & ECCCo (no CP) & 1.39 ± 0.08** & 0.69 ± 0.05** & 0.11 ± 0.00** & 1.21 ± 0.07*\hphantom{*} & 0.77 ± 0.07** & 0.39 ± 0.01\hphantom{*}\hphantom{*}\\

    & ECCCo (no EBM) & 1.70 ± 0.09\hphantom{*}\hphantom{*} & 0.99 ± 0.08\hphantom{*}\hphantom{*} & 0.14 ± 0.00*\hphantom{*} & 1.31 ± 0.07\hphantom{*}\hphantom{*} & 0.97 ± 0.10\hphantom{*}\hphantom{*} & 0.32 ± 0.01**\\

    & REVISE & 1.39 ± 0.15** & \textbf{0.59 ± 0.04}** & 0.25 ± 0.07\hphantom{*}\hphantom{*} & 1.01 ± 0.07** & \textbf{0.63 ± 0.04}** & 0.33 ± 0.07\hphantom{*}\hphantom{*}\\

    & Schut & 1.59 ± 0.10*\hphantom{*} & 1.10 ± 0.06\hphantom{*}\hphantom{*} & \textbf{0.09 ± 0.00}** & 1.34 ± 0.07\hphantom{*}\hphantom{*} & 1.21 ± 0.10\hphantom{*}\hphantom{*} & \textbf{0.26 ± 0.01}**\\

\multirow{-7}{*}{\raggedright\arraybackslash JEM Ensemble} & Wachter & 1.71 ± 0.09\hphantom{*}\hphantom{*} & 0.99 ± 0.08\hphantom{*}\hphantom{*} & 0.14 ± 0.00\hphantom{*}\hphantom{*} & 1.31 ± 0.08\hphantom{*}\hphantom{*} & 0.95 ± 0.10\hphantom{*}\hphantom{*} & 0.33 ± 0.01\hphantom{*}\hphantom{*}\\
\bottomrule
\end{tabular}}
\caption{Results for tabular datasets: sample averages +/- one standard deviation across valid counterfactuals. The best outcomes are highlighted in bold. Asterisks indicate that the given value is more than one (*) or two (**) standard deviations away from the baseline (\textit{Wachter}). \label{tab:results-tabular}}
\end{table*}

Overall, we find strong empirical evidence suggesting that \textit{ECCCo} consistently achieves state-of-the-art faithfulness. Across all models and datasets highlighted here, different variations of \textit{ECCCo} consistently outperform other generators with respect to faithfulness, in many cases substantially. This pattern is mostly robust across all other datasets. 

In particular, we note that the best results are generally obtained when using the full \textit{ECCCo} objective (Equation~\ref{eq:eccco}). In other words, constraining both energy and predictive uncertainty typically yields the most faithful counterfactuals. We expected the former to play a more significant role in this context and that is typically what we find across all datasets. The results in Table~\ref{tab:results-tabular} indicate that faithfulness can be improved substantially by relying solely on the energy constraint (\textit{ECCCo (no CP)}). In most cases, however, the full objective yields the most faithful counterfactuals. This indicates that predictive uncertainty minimization plays an important role in achieving faithfulness. 

We also generally find that latent space search does not impede faithfulness for \textit{ECCCo}. In most cases \textit{ECCCo+} is either on par with \textit{ECCCo} or even outperforms it. There are some notable exceptions though. Cases in which \textit{ECCCo} achieves substantially better faithfulness without latent space search tend to involve more vulnerable models like the simple MLP for MNIST in Table~\ref{tab:results-vision}. We explain this finding as follows: even though dimensionality reduction through PCA in the case of \textit{ECCCo+} can be considered a relatively mild form of intervention, the first $n_z$ principal components fail to capture some of the variation in the data. More vulnerable models may be particularly sensitive to this residual variation in the data. 

Consistent with this finding, we also observe that \textit{REVISE} ranks higher for faithfulness, if the model itself has learned more plausible representations of the underlying data: \textit{REVISE} generates more faithful counterfactuals than the baseline for the \textit{JEM} Ensemble in Table~\ref{tab:results-tabular} and the LeNet-5 \textit{CNN} in Table~\ref{tab:results-vision}. This demonstrates that the two desiderata---faithfulness and plausibility---are not mutually exclusive.

\subsection{Balancing Desiderata}

\begin{figure}
  \centering
  \includegraphics[width=1.0\linewidth]{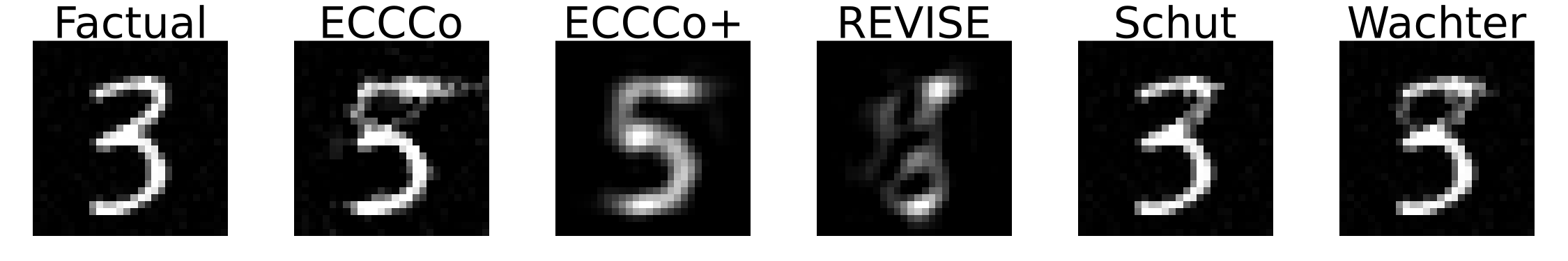}
  \caption{Counterfactuals for turning a 3 into a 5: factual (left), then the counterfactuals generated by \textit{ECCCo}, \textit{ECCCo+}, \textit{REVISE}, \textit{Schut} and \textit{Wachter}.}\label{fig:mnist-bmk}
\end{figure}

\begin{table}
\centering
\resizebox{\linewidth}{!}{
\begin{tabular}[t]{llcc}
\toprule
\multicolumn{2}{c}{ } & \multicolumn{2}{c}{MNIST} \\
\cmidrule(l{3pt}r{3pt}){3-4}
Model & Generator & Unfaithfulness ↓ & Implausibility ↓\\
\midrule
    & ECCCo & \textbf{0.243 ± 0.000}** & 0.420 ± 0.001\hphantom{*}\hphantom{*}\\

    & ECCCo+ & 0.246 ± 0.000*\hphantom{*} & 0.306 ± 0.001**\\

    & REVISE & 0.248 ± 0.000\hphantom{*}\hphantom{*} & \textbf{0.301 ± 0.004}**\\

    & Schut & 0.247 ± 0.001\hphantom{*}\hphantom{*} & 0.303 ± 0.008**\\

\multirow{-5}{*}{\raggedright\arraybackslash MLP} & Wachter & 0.247 ± 0.000\hphantom{*}\hphantom{*} & 0.344 ± 0.002\hphantom{*}\hphantom{*}\\
\cmidrule{1-4}
    & ECCCo & 0.248 ± 0.000** & 0.387 ± 0.002\hphantom{*}\hphantom{*}\\

    & ECCCo+ & \textbf{0.248 ± 0.000}** & 0.310 ± 0.002**\\

    & REVISE & 0.248 ± 0.000** & 0.301 ± 0.002**\\

    & Schut & 0.250 ± 0.002\hphantom{*}\hphantom{*} & \textbf{0.289 ± 0.024}*\hphantom{*}\\

\multirow{-5}{*}{\raggedright\arraybackslash LeNet-5} & Wachter & 0.249 ± 0.000\hphantom{*}\hphantom{*} & 0.335 ± 0.002\hphantom{*}\hphantom{*}\\
\bottomrule
\end{tabular}}
\caption{Results for vision dataset. Formatting details are the same as in Table~\ref{tab:results-tabular}. \label{tab:results-vision}}
\end{table}

Overall, we find strong empirical evidence suggesting that \textit{ECCCo} can achieve near state-of-the-art plausibility without sacrificing faithfulness. Figure~\ref{fig:mnist-bmk} shows one such example taken from the \textit{MNIST} benchmark where the objective is to turn the factual `three' (far left) into a `five'. The underlying model is a LeNet-5 \textit{CNN}. The different images show the counterfactuals produced by the generators, of which all but the one produced by \textit{Schut} are valid. Both variations of \textit{ECCCo} produce plausible counterfactuals.

Looking at the benchmark results presented in Tables~\ref{tab:results-tabular} and~\ref{tab:results-vision} we firstly note that although \textit{REVISE} generally performs best, \textit{ECCCo} and in particular \textit{ECCCo+} often approach SOTA performance. Upon visual inspection of the generated images we actually find that \textit{ECCCo+} performs much better than \textit{REVISE} (see appendix). Zooming in on the details we observe that \textit{ECCCo} and its variations do particularly well, whenever the underlying model has been explicitly trained to learn plausible representations of the data. For both tabular datasets in Table~\ref{tab:results-tabular}, \textit{ECCCo} improves plausibility substantially compared to the baseline. This broad pattern is mostly consistent for all other datasets, although there are notable exceptions for which \textit{ECCCo} takes the lead on both plausibility and faithfulness. 

While we maintain that generally speaking plausibility should hinge on the quality of the model, our results also indicate that it is possible to balance faithfulness and plausibility if needed: \textit{ECCCo+} generally outperforms other variants of \textit{ECCCo} in this context, occasionally at the small cost of slightly reduced faithfulness. For the vision datasets especially, we find that  \textit{ECCCo+} is consistently second only to \textit{REVISE} for all models and regularly substantially better than the baseline. Looking at the \textit{California Housing} data, latent space search markedly improves plausibility without sacrificing faithfulness: for the \textit{JEM} Ensemble, \textit{ECCCo+} performs substantially better than the baseline and only marginally worse than \textit{REVISE}. Importantly, \textit{ECCCo+} does not attain plausibility at all costs: for the \textit{MLP} Ensemble, plausibility is still very low but this seems to faithfully represent what the model has learned. 

We conclude from the findings presented thus far that \textit{ECCCo} enables us to reconcile the objectives of faithfulness and plausibility. It produces plausible counterfactuals if and only if the model itself has learned plausible explanations for the data. It thus avoids the risk of generating plausible but potentially misleading explanations for models that are highly susceptible to implausible explanations.

\subsection{Additional Desiderata}

While we have deliberately focused on our key metrics of interest so far, it is worth briefly considering other common desiderata for counterfactuals. With reference to the right-most columns for each dataset in Table~\ref{tab:results-tabular}, we firstly note that \textit{ECCCo} typically reduces predictive uncertainty as intended. Consistent with its design, \textit{Schut} performs well on this metric even though it does not explicitly address uncertainty as measured by conformal prediction set sizes. 

Another commonly discussed desideratum is closeness~\citep{wachter2017counterfactual}: counterfactuals that are closer to their factuals are associated with smaller costs to individuals in the context of algorithmic recourse. As evident from the additional tables in the appendix, the closeness desideratum tends to be negatively correlated with plausibility and faithfulness. Consequently, both \textit{REVISE} and \textit{ECCCo} generally yield more costly counterfactuals than the baseline. Nonetheless, \textit{ECCCo} does not seem to stretch costs unnecessarily: in Figure~\ref{fig:mnist-bmk} useful parts of the factual `three' are clearly retained.

\section{Limitations}

Despite having taken considerable measures to study our methodology carefully, limitations can still be identified. 

Firstly, we recognize that our proposed distance-based evaluation metrics for plausibility and faithfulness may not be universally applicable to all types of data. In any case, they depend on choosing a distance metric on a case-by-case basis, as we have done in this work. Arguably, commonly used metrics for measuring other desiderata such as closeness suffer from the same pitfall. We therefore think that future work on counterfactual explanations could benefit from defining universal evaluation metrics. 

Relatedly, we note that our proposed metric for measuring faithfulness depends on the availability of samples generated through SGLD, which in turn requires gradient access for models. This means it cannot be used to evaluate non-differentiable classifiers. Consequently, we also have not applied \textit{ECCCo} to some machine learning models commonly used for classification such as decision trees. Since \textit{ECCCo} itself does not rely on SGLD, its defining penalty functions are indeed applicable to gradient-free counterfactual generators. This is an interesting avenue for future research.

Next, common challenges associated with energy-based modelling including sensitivity to scale, training instabilities and sensitivity to hyperparameters also apply to \textit{ECCCo} to some extent. In grid searches for optimal hyperparameters, we have noticed that unless properly regularized, \textit{ECCCo} is sometimes prone to overshoot for the energy constraint. 

Finally, while we have used ablation to understand the roles of the different components of \textit{ECCCo}, the scope of this work has prevented us from investigating the role of conformal prediction in this context more thoroughly. We have exclusively relied on split conformal prediction and have used fixed values for the predetermined error rate and other hyperparameters. Future work could benefit from more extensive ablation studies that tune hyperparameters and investigate different approaches to conformal prediction.

\section{Conclusion}

This work leverages ideas from energy-based modelling and conformal prediction in the context of counterfactual explanations. We have proposed a new way to generate counterfactuals that are maximally faithful to the black-box model they aim to explain. Our proposed generator, \textit{ECCCo}, produces plausible counterfactuals iff the black-box model itself has learned realistic explanations for the data, which we have demonstrated through rigorous empirical analysis. This should enable researchers and practitioners to use counterfactuals in order to discern trustworthy models from unreliable ones. While the scope of this work limits its generalizability, we believe that \textit{ECCCo} offers a solid base for future work on faithful counterfactual explanations.

\section*{Acknowledgements}

Some of the members of TU Delft were partially funded by ICAI AI for Fintech Research, an ING---TU Delft
collaboration. 

Research reported in this work was partially or completely facilitated by computational resources and support of the DelftBlue~\citep{DHPC2022} and the Delft AI Cluster (DAIC: https://doc.daic.tudelft.nl/) at TU Delft. Detailed information about the utilized computing resources can be found in the appendix. The authors would like to thank Azza Ahmed, in particular, for her tremendous help with running Julia jobs on the cluster. The work remains the sole responsibility of the authors.

We would also like to express our gratitude to the group of students who have recently contributed to the development of CounterfactualExplanations.jl~\citep{altmeyer2023explaining}, the Julia package that was used for this analysis: Rauno Arike, Simon Kasdorp, Lauri Kesküll, Mariusz Kicior, Vincent Pikand.

All code used for the analysis in this paper can be found here: https://github.com/pat-alt/ECCCo.jl.

%% file: partials/appendix.tex
\appendix
\section*{Appendices}
\renewcommand{\thesubsection}{\Alph{subsection}}

The following appendices provide additional details that are relevant to the paper. Appendices~\ref{app:jem} and~\ref{app:cp} explain any tasks related to Energy-Based Modelling and Predictive Uncertainty Quantification through Conformal Prediction, respectively. Appendix~\ref{app:eccco} provides additional technical and implementation details about our proposed generator, \textit{ECCCo}, including references to our open-sourced code base. A complete overview of our experimental setup detailing our parameter choices, training procedures and initial black-box model performance can be found in Appendix~\ref{app:setup}. Finally, Appendix~\ref{app:results} reports all of our experimental results in more detail.

\subsection{Energy-Based Modelling}\label{app:jem}

Since we were not able to identify any existing open-source software for Energy-Based Modelling that would be flexible enough to cater to our needs, we have developed a \texttt{Julia} package from scratch. The package has been open-sourced, but to avoid compromising the double-blind review process, we refrain from providing more information at this stage. In our development we have heavily drawn on the existing literature:~\citet{du2019implicit} describe best practices for using EBM for generative modelling;~\citet{grathwohl2020your} explain how EBM can be used to train classifiers jointly for the discriminative and generative tasks. We have used the same package for training and inference, but there are some important differences between the two cases that are worth highlighting here.

\subsubsection{Training: Joint Energy Models}

To train our Joint Energy Models we broadly follow the approach outlined in~\citet{grathwohl2020your}. Formally, JEMs are defined by the following joint distribution:

\begin{equation}
  \begin{aligned}
    \log p_{\theta}(\mathbf{x},\mathbf{y}) &= \log p_{\theta}(\mathbf{y}|\mathbf{x}) + \log p_{\theta}(\mathbf{x})
  \end{aligned}
\end{equation}

Training therefore involves a standard classification loss component $L_{\text{clf}}(\theta)=-\log p_{\theta}(\mathbf{y}|\mathbf{x})$ (e.g. cross-entropy loss) as well as a generative loss component $L_{\text{gen}}(\theta)=-\log p_{\theta}(\mathbf{x})$. Analogous to how we defined the conditional distribution over inputs in Definition~\ref{def:faithful}, $p_{\theta}(\mathbf{x})$ denotes the unconditional distribution over inputs. The model gradient of this component of the loss function can be expressed as follows:

\begin{equation}\label{eq:gen-true}
  \begin{aligned}
    \nabla_{\theta}L_{\text{gen}}(\theta)&=-\nabla_{\theta}\log p_{\theta}(\mathbf{x})=-\left(\mathbb{E}_{p(\mathbf{x})} \left\{  \nabla_{\theta} \mathcal{E}_{\theta}(\mathbf{x}) \right\} - \mathbb{E}_{p_{\theta}(\mathbf{x})} \left\{  \nabla_{\theta} \mathcal{E}_{\theta}(\mathbf{x}) \right\} \right)
  \end{aligned}
\end{equation}

To draw samples from $p_{\theta}(\mathbf{x})$, we rely exclusively on the conditional sampling approach described in~\citet{grathwohl2020your} for both training and inference: we first draw $\mathbf{y}\sim p(\mathbf{y})$ and then sample $\mathbf{x} \sim p_{\theta}(\mathbf{x}|\mathbf{y})$~\citep{grathwohl2020your} via Equation~\ref{eq:sgld} with energy $\mathcal{E}_{\theta}(\mathbf{x}|\mathbf{y})=\mu_{\theta}(\mathbf{x})[\mathbf{y}]$ where $\mu_{\theta}: \mathcal{X} \mapsto \mathbb{R}^K$ returns the linear predictions (logits) of our classifier $M_{\theta}$. While our package also supports unconditional sampling, we found conditional sampling to work well. It is also well aligned with CE, since in this context we are interested in conditioning on the target class. 

As mentioned in the body of the paper, we rely on a biased sampler involving separately specified values for the step size $\epsilon$ and the standard deviation $\sigma$ of the stochastic term involving $\mathbf{r}$. Formally, our biased sampler performs updates as follows: 

\begin{equation}\label{eq:biased-sgld}
  \begin{aligned}
    \hat{\mathbf{x}}_{j+1} &\leftarrow \hat{\mathbf{x}}_j - \frac{\phi}{2} \mathcal{E}_{\theta}(\hat{\mathbf{x}}_j|\mathbf{y}^+) + \sigma \mathbf{r}_j, && j=1,...,J
  \end{aligned}
\end{equation}

Consistent with~\citet{grathwohl2020your}, we have specified $\phi=2$ and $\sigma=0.01$ as the default values for all of our experiments. Here we have deliberately departed slightly from the notation in Equation~\ref{eq:sgld} to emphasize that we use fixed values for $\phi$ and $\sigma$, consistent with the related literature. The number of total SGLD steps $J$ varies by dataset (Table~\ref{tab:ebmparams}). Following best practices, we initialize $\mathbf{x}_0$ randomly in 5\% of all cases and sample from a buffer in all other cases. The buffer itself is randomly initialised and gradually grows to a maximum of 10,000 samples during training as $\hat{\mathbf{x}}_{J}$ is stored in each epoch~\citep{du2019implicit,grathwohl2020your}. 

It is important to realise that sampling is done during each training epoch, which makes training Joint Energy Models significantly harder than conventional neural classifiers. In each epoch the generated (batch of) sample(s) $\hat{\mathbf{x}}_{J}$ is used as part of the generative loss component, which compares its energy to that of observed samples $\mathbf{x}$: 

\begin{equation}\label{eq:gen-loss}
  \begin{aligned}
    L_{\text{gen}}(\theta)&\approx\mu_{\theta}(\mathbf{x})[\mathbf{y}]-\mu_{\theta}(\hat{\mathbf{x}}_{J})[\mathbf{y}]
  \end{aligned}
\end{equation}

Our full training objective can be summarized as follows,

\begin{equation}\label{eq:jem-loss}
  \begin{aligned}
    L_{\text{JEM}}(\theta) &= L_{\text{clf}}(\theta) + L_{\text{gen}}(\theta) + \lambda L_{\text{reg}}(\theta) 
  \end{aligned}
\end{equation}

where $L_{\text{reg}}(\theta)$ is a Ridge penalty (L2 norm) that regularises energy magnitudes for both observed and generated samples~\citep{du2019implicit}. We have used varying degrees of regularization depending on the dataset ($\lambda$ in Table~\ref{tab:ebmparams}). 

\begin{figure}
  \centering
  \includegraphics[width=0.75\linewidth]{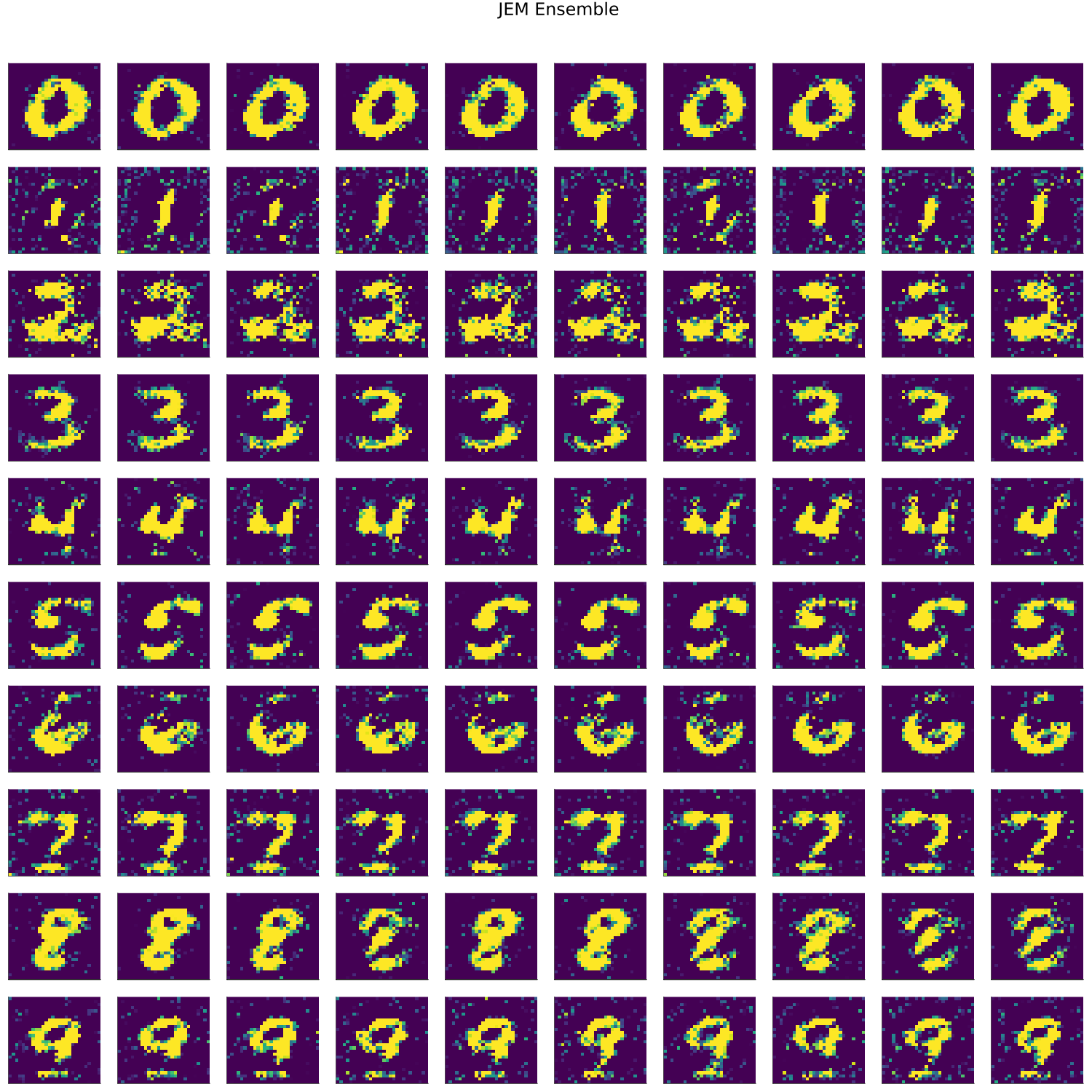}
  \caption{Conditionally generated \textit{MNIST} images for our JEM Ensemble.}\label{fig:mnist-gen}
\end{figure}

\begin{figure}
  \centering
  \includegraphics[width=0.5\linewidth]{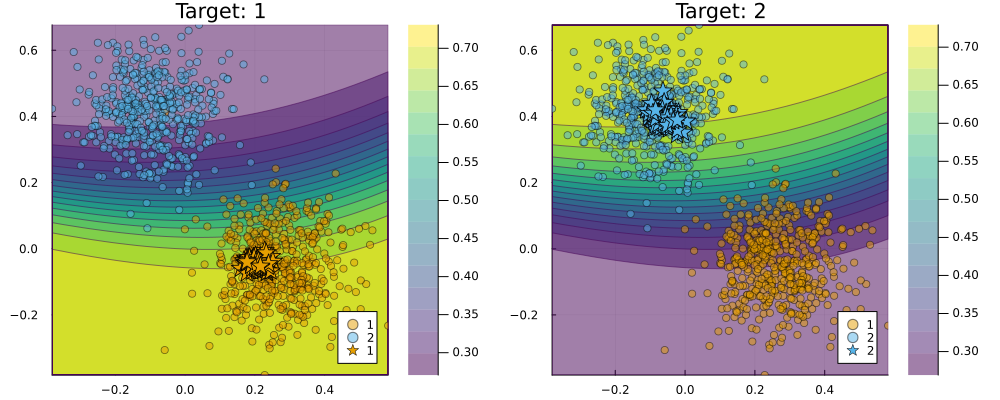}
  \caption{Conditionally generated samples (stars) for our \textit{Linearly Separable} data using a JEM.}\label{fig:poc-gen}
\end{figure}

Contrary to existing work, we have not typically used the entire minibatch of training data for the generative loss component but found that using a subset of the minibatch was often sufficient in attaining decent generative performance (Table~\ref{tab:ebmparams}). This has helped to reduce the computational burden for our models, which should make it easier for others to reproduce our findings. Figures~\ref{fig:mnist-gen} and~\ref{fig:poc-gen} show generated samples for our \textit{MNIST} and \textit{Moons} data, to provide a sense of their generative property.

\import{../contents/}{table_ebm_params.tex}

\subsubsection{Inference: Quantifying Models' Generative Property}

At inference time, we assume no prior knowledge about the model's generative property. This means that we do not tab into the existing buffer of generated samples for our Joint Energy Models, but instead generate conditional samples from scratch. While we have relied on the default values $\epsilon=2$ and $\sigma=0.01$ also during inference, the number of total SGLD steps was set to $J=500$ in all cases, so significantly higher than during training. For all of our synthetic datasets and models, we generated 50 conditional samples and then formed subsets containing the $n_{E}=25$ lowest-energy samples. While in practice it would be sufficient to do this once for each model and dataset, we have chosen to perform sampling separately for each individual counterfactual in our experiments to account for stochasticity. To help reduce the computational burden for our real-world datasets we have generated only 10 conditional samples each time and used all of them in our counterfactual search. Using more samples, as we originally did, had no substantial impact on our results.

\subsection{Conformal Prediction}\label{app:cp}

In this Appendix~\ref{app:cp} we provide some more background on CP and explain in some more detail how we have used recent advances in Conformal Training for our purposes.

\subsubsection{Background on CP}

Intuitively, CP works under the premise of turning heuristic notions of uncertainty into rigorous uncertainty estimates by repeatedly sifting through the data. It can be used to generate prediction intervals for regression models and prediction sets for classification models. Since the literature on CE and AR is typically concerned with classification problems, we focus on the latter. A particular variant of CP called Split Conformal Prediction (SCP) is well-suited for our purposes, because it imposes only minimal restrictions on model training. 

Specifically, SCP involves splitting the data $\mathcal{D}_n=\{(\mathbf{x}_i,\mathbf{y}_i)\}_{i=1,...,n}$ into a proper training set $\mathcal{D}_{\text{train}}$ and a calibration set $\mathcal{D}_{\text{cal}}$. The former is used to train the classifier in any conventional fashion. The latter is then used to compute so-called nonconformity scores: $\mathcal{S}=\{s(\mathbf{x}_i,\mathbf{y}_i)\}_{i \in \mathcal{D}_{\text{cal}}}$ where $s: (\mathcal{X},\mathcal{Y}) \mapsto \mathbb{R}$ is referred to as \textit{score function}. In the context of classification, a common choice for the score function is just $s_i=1-M_{\theta}(\mathbf{x}_i)[\mathbf{y}_i]$, that is one minus the softmax output corresponding to the observed label $\mathbf{y}_i$~\citep{angelopoulos2021gentle}. 

Finally, classification sets are formed as follows,

\begin{equation}\label{eq:scp}
  \begin{aligned}
    C_{\theta}(\mathbf{x}_i;\alpha)=\{\mathbf{y}: s(\mathbf{x}_i,\mathbf{y}) \le \hat{q}\}
  \end{aligned}
\end{equation}

where $\hat{q}$ denotes the $(1-\alpha)$-quantile of $\mathcal{S}$ and $\alpha$ is a predetermined error rate. As the size of the calibration set increases, the probability that the classification set $C(\mathbf{x}_{\text{test}})$ for a newly arrived sample $\mathbf{x}_{\text{test}}$ does not cover the true test label $\mathbf{y}_{\text{test}}$ approaches $\alpha$~\citep{angelopoulos2021gentle}. 

Observe from Equation~\ref{eq:scp} that Conformal Prediction works on an instance-level basis, much like CE are local. The prediction set for an individual instance $\mathbf{x}_i$ depends only on the characteristics of that sample and the specified error rate. Intuitively, the set is more likely to include multiple labels for samples that are difficult to classify, so the set size is indicative of predictive uncertainty. To see why this effect is exacerbated by small choices for $\alpha$ consider the case of $\alpha=0$, which requires that the true label is covered by the prediction set with probability equal to 1.

\subsubsection{Differentiability}\label{app:cp-diff}

The fact that conformal classifiers produce set-valued predictions introduces a challenge: it is not immediately obvious how to use such classifiers in the context of gradient-based counterfactual search. Put differently, it is not clear how to use prediction sets in Equation~\ref{eq:general}. Fortunately, \citet{stutz2022learning} have recently proposed a framework for Conformal Training that also hinges on differentiability. Specifically, they show how Stochastic Gradient Descent can be used to train classifiers not only for the discriminative task but also for additional objectives related to Conformal Prediction. One such objective is \textit{efficiency}: for a given target error rate $\alpha$, the efficiency of a conformal classifier improves as its average prediction set size decreases. To this end, the authors introduce a smooth set size penalty defined in Equation~\ref{eq:setsize} in the body of this paper. Formally, it is defined as $C_{\theta,\mathbf{y}}(\mathbf{x}_i;\alpha):=\sigma\left((s(\mathbf{x}_i,\mathbf{y})-\alpha) T^{-1}\right)$ for $\mathbf{y}\in\mathcal{Y}$, where $\sigma$ is the sigmoid function and $T$ is a hyper-parameter used for temperature scaling~\citep{stutz2022learning}.

\begin{figure}
  \centering
  \includegraphics[width=1.0\linewidth]{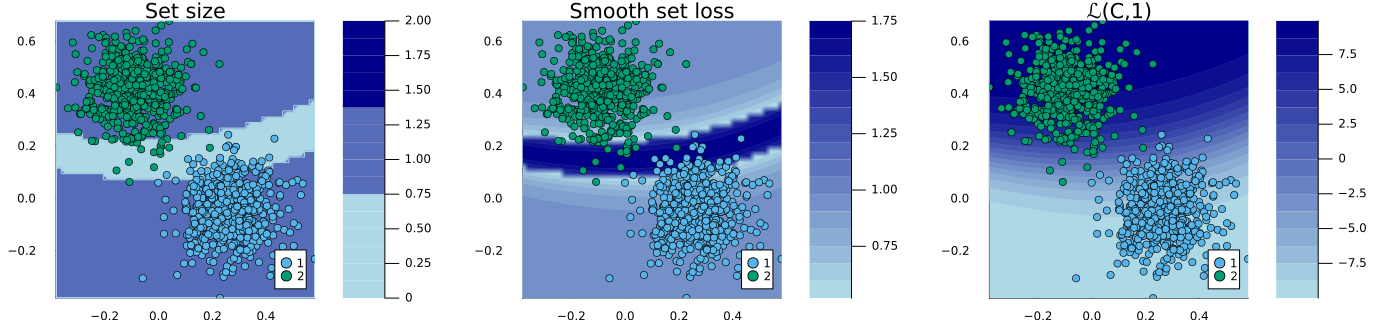}
  \caption{Prediction set size (left), smooth set size loss (centre) and configurable classification loss (right) for a JEM trained on our \textit{Linearly Separable} data.}\label{fig:cp-diff}
\end{figure}

In addition to the smooth set size penalty,~\citet{stutz2022learning} also propose a configurable classification loss function, that can be used to enforce coverage. For \textit{MNIST} data, we found that using this function generally improved the visual quality of the generated counterfactuals, so we used it in our experiments involving real-world data. For the synthetic dataset, visual inspection of the counterfactuals showed that using the configurable loss function sometimes led to overshooting: counterfactuals would end up deep inside the target domain but far away from the observed samples. For this reason, we instead relied on standard cross-entropy loss for our synthetic datasets. As we have noted in the body of the paper, more experimental work is certainly needed in this context. Figure~\ref{fig:cp-diff} shows the prediction set size (left), smooth set size loss (centre) and configurable classification loss (right) for a \textit{JEM} trained on our \textit{Linearly Separable} data.

\subsection{\textit{ECCCo}}\label{app:eccco}

In this section, we explain \textit{ECCCo} in some more detail, briefly discuss convergence conditions for counterfactual explanations and provide details concerning the actual implementation of our framework in \texttt{Julia}.  

\subsubsection{Deriving the search objective} 

The counterfactual search objective for \textit{ECCCo} was introduced in Equation~\ref{eq:eccco} in the body of the paper. We restate this equation here for reference:

\begin{equation} \label{eq:eccco-app}
  \begin{aligned}
  \mathbf{Z}^\prime &= \arg \min_{\mathbf{Z}^\prime \in \mathcal{Z}^L} \{  {\text{yloss}(M_{\theta}(f(\mathbf{Z}^\prime)),\mathbf{y}^+)}+ \lambda_{1} {\text{dist}(f(\mathbf{Z}^\prime),\mathbf{x}) } \\
  &+ \lambda_2 \mathcal{E}_{\theta}(\mathbf{Z}^\prime,\widehat{\mathbf{X}}_{\theta,\mathbf{y}^+}) + \lambda_3 \Omega(C_{\theta}(f(\mathbf{Z}^\prime);\alpha)) \} 
  \end{aligned} 
\end{equation}

We can make the connection to energy-based modeling more explicit by restating the counterfactual search objective in terms $L_{\text{JEM}}(\theta)$, which we defined in Equation~\ref{eq:jem-loss}. In particular, consider the following counterfactual search objective,

\begin{equation} \label{eq:eccco-jem}
  \begin{aligned}
  \mathbf{Z}^\prime &= \arg \min_{\mathbf{Z}^\prime \in \mathcal{Z}^L} \{  {L_{\text{JEM}}(\theta;M_{\theta}(f(\mathbf{Z}^\prime)),\mathbf{y}^+)}+ \lambda_{1} {\text{dist}(f(\mathbf{Z}^\prime),\mathbf{x}) }  + \lambda_3 \Omega(C_{\theta}(f(\mathbf{Z}^\prime);\alpha)) \} 
  \end{aligned} 
\end{equation}

where we have simply used the JEM loss function as $\text{yloss}(M_{\theta}(f(\mathbf{Z}^\prime)),\mathbf{y}^+)$.

Now note that aside from the additional penalties in Equation~\ref{eq:eccco-app}, the only key difference between our counterfactual search objective and the joint-energy training objective is the parameter that is being optimized. In joint-energy training we optimize the objective with respect to the network weights $\theta$. Recall that $\mathcal{E}_{\theta}(\mathbf{x}|\mathbf{y})=\mu_{\theta}(\mathbf{x})[\mathbf{y}]$. Then the partial gradient with respect to the generative loss component of $L_{\text{JEM}}(\theta)$ can be expressed as follows:

\begin{equation}\label{eq:jem-grad}
  \begin{aligned}
    \nabla_{\theta}L_{\text{gen}}(\theta) &= \nabla_{\theta}\mu_{\theta}(\mathbf{x})[\mathbf{y}]- \nabla_{\theta}\mu_{\theta}(\hat{\mathbf{x}}_{J})[\mathbf{y}]
  \end{aligned}
\end{equation}

During the counterfactual search, we take the network parameters as fixed and instead optimize with respect to the counterfactual itself\footnote{Here we omit the notion of a latent search space to make the comparison easier.},

\begin{equation}\label{eq:ce-grad}
  \begin{aligned}
    \nabla_{\mathbf{x}}L_{\text{gen}}(\theta) &= \nabla_{\mathbf{x}}\mu_{\theta}(\mathbf{x})[\mathbf{y}^+]- \nabla_{\mathbf{x}}\mu_{\theta}(\hat{\mathbf{x}}_{J})[\mathbf{y}^+]=\nabla_{\mathbf{x}}\mu_{\theta}(\mathbf{x})[\mathbf{y}^+]=\nabla_{\mathbf{x}}\mathcal{E}_{\theta}(\mathbf{x}|\mathbf{y}^+)
  \end{aligned}
\end{equation}

where the second term is equal to zero because $\mu_{\theta}(\hat{\mathbf{x}}_{J})[\mathbf{y}]$ is invariant with respect to $\mathbf{x}$. Since this term has zero gradients, we can remove it from the loss function altogether. For the regularization loss component of $L_{\text{JEM}}(\theta)$ we can proceed analogously such that we can rewrite Equation~\ref{eq:eccco-jem} as follows:

\begin{equation} \label{eq:eccco-jem-2}
  \begin{aligned}
  \mathbf{Z}^\prime =& \arg \min_{\mathbf{Z}^\prime \in \mathcal{Z}^L} \{  {\text{yloss}(M_{\theta}(f(\mathbf{Z}^\prime)),\mathbf{y}^+) + \mathcal{E}_{\theta}(f(\mathbf{Z}^\prime)|\mathbf{y}^+) + || \mathcal{E}_{\theta}(f(\mathbf{Z}^\prime)|\mathbf{y}^+) ||_2^2} \\ &+ \lambda_{1} {\text{dist}(f(\mathbf{Z}^\prime),\mathbf{x}) }  + \lambda_3 \Omega(C_{\theta}(f(\mathbf{Z}^\prime);\alpha)) \} 
  \end{aligned} 
\end{equation}

Now we notice that Equation~\ref{eq:eccco-jem-2} is equivalent to Equation~\ref{eq:eccco-app} for $\lambda_2=1$. For the sake of simplicity, we omitted the regularization component from Equation~\ref{eq:eccco} in the main text. Intuitively, taking iterative gradient steps according to Equation~\ref{eq:ce-grad} has the effect of constraining the energy of the counterfactual until. The generative property of the underlying model implicitly enters this equation through $\theta$.

\subsubsection{The \textit{ECCCo} algorithm}

Algorithm~\ref{alg:eccco} describes how exactly \textit{ECCCo} works. For the sake of simplicity and without loss of generality, we limit our attention to generating a single counterfactual $\mathbf{x}^\prime=f(\mathbf{z}^\prime)$. The counterfactual state $\mathbf{z}^\prime$ is initialized at the factual $\mathbf{x}$. Other forms of initialization are also suitable but not considered here. For example, one may choose at a small random perturbation to all features~\citep{slack2021counterfactual}. Next, we calibrate the model $M_{\theta}$ through split conformal prediction. Finally, we search counterfactuals through gradient descent where $\mathcal{L}(\mathbf{z}^\prime,\mathbf{y}^+,\widehat{\mathbf{X}}_{\theta,\mathbf{y}^+}; \Lambda, \alpha)$ denotes our loss function defined in Equation~\ref{eq:eccco}. The search terminates once the convergence criterium is met or the maximum number of iterations $T$ has been exhausted. Note that the choice of convergence criterium has important implications on the final counterfactual which we explain below.

\begin{algorithm*}[h]
  \caption{The \textit{ECCCo} generator}\label{alg:eccco}
  \begin{algorithmic}[1]
    \Require $\mathbf{x}, \mathbf{y}^+, M_{\theta}, \Lambda=[\lambda_1,\lambda_2,\lambda_3], \alpha, \mathcal{D}, T$ where $M_{\theta}(\mathbf{x})\neq\mathbf{y}^+$
    \Ensure $\mathbf{x}^\prime$
    \State Initialize $\mathbf{z}^\prime \gets \mathbf{x}$ 
    \State Run \textit{SCP} for $M_{\theta}$ using $\mathcal{D}$ \Comment{Calibrate model through split conformal prediction.}
    \State Initialize $t \gets 0$
    \While{\textit{not converged} or $t < T$} \Comment{For convergence conditions see below.}
    \State $\mathbf{z}^\prime \gets \mathbf{z}^\prime - \eta \nabla_{\mathbf{z}^\prime} \mathcal{L}(\mathbf{z}^\prime,\mathbf{y}^+; \Lambda, \alpha)$ \Comment{Take gradient step of size $\eta$.}
    \State $t \gets t+1$
    \EndWhile
    \State $\mathbf{x}^\prime \gets \mathbf{z}^\prime$
  \end{algorithmic}
\end{algorithm*}

\subsubsection{The \textit{ECCCo+} algorithm}

Algorithm~\ref{alg:eccco-plus} describes how exactly \textit{ECCCo+} works. The only difference to \textit{ECCCo} is that we encode and decode features using PCA. In particular, we let $f^{-1}(\mathbf{x})$ denote the projection of $\mathbf{x}$ to its first $n_z$ principal components. Conversely, $f(\mathbf{z}^\prime)$ maps back from the projection to the feature space. 

\begin{algorithm*}[h]
  \caption{The \textit{ECCCo+} generator}\label{alg:eccco-plus}
  \begin{algorithmic}[1]
    \Require $\mathbf{x}, \mathbf{y}^+, M_{\theta}, f, \Lambda=[\lambda_1,\lambda_2,\lambda_3], \alpha, \mathcal{D}, T$ where $M_{\theta}(\mathbf{x})\neq\mathbf{y}^+$
    \Ensure $\mathbf{x}^\prime$
    \State Initialize $\mathbf{z}^\prime \gets f^{-1}(\mathbf{x})$ \Comment{Map to counterfactual state space.}
    \State Run \textit{SCP} for $M_{\theta}$ using $\mathcal{D}$ \Comment{Calibrate model through split conformal prediction.}
    \State Initialize $t \gets 0$
    \While{\textit{not converged} or $t < T$} \Comment{For convergence conditions see below.}
    \State $\mathbf{z}^\prime \gets \mathbf{z}^\prime - \eta \nabla_{\mathbf{z}^\prime} \mathcal{L}(\mathbf{z}^\prime,\mathbf{y}^+; \Lambda, \alpha)$ \Comment{Take gradient step of size $\eta$.}
    \State $t \gets t+1$
    \EndWhile
    \State $\mathbf{x}^\prime \gets f(\mathbf{z}^\prime)$ \Comment{Map back to feature space.}
  \end{algorithmic}
\end{algorithm*}

\subsubsection{The \textit{ECCCo-L1} algorithm}

Algorithm~\ref{alg:eccco-l1} describes a variation of \textit{ECCCo} that we initally considered but ultimately discarded. For the sake of completeness we have included this approach here in the appendix. It generally yields very faithful counterfactuals but it is computationally much more expensive and struggles with plausibility.

Instead of constraining energy directly, this approach works under the premise of penalizing the distance between the counterfactual and samples generated through SGLD. The counterfactual state $\mathbf{z}^\prime$ is initialized as in Algorithm~\ref{alg:eccco}. Next, we generate $n_{\mathcal{B}}$ conditional samples $\hat{\mathbf{x}}_{\theta,\mathbf{y}^+}$ using SGLD (Equation~\ref{eq:sgld}) and store the $n_E$ instances with the lowest energy. We then calibrate the model $M_{\theta}$ through split conformal prediction. Finally, we search counterfactuals through gradient descent where $\mathcal{L}(\mathbf{z}^\prime,\mathbf{y}^+,\widehat{\mathbf{X}}_{\theta,\mathbf{y}^+}; \Lambda, \alpha)$ denotes our loss function defined in Equation~\ref{eq:eccco}, but instead of constraining energy directly we use Equaqtion~\ref{eq:faith} (unfaithfulness metric) as a penalty.

\begin{algorithm*}[h]
  \caption{The \textit{ECCCo-L1} generator}\label{alg:eccco-l1}
  \begin{algorithmic}[1]
    \Require $\mathbf{x}, \mathbf{y}^+, M_{\theta}, f, \Lambda=[\lambda_1,\lambda_2,\lambda_3], \alpha, \mathcal{D}, T, \eta, n_{\mathcal{B}}, n_E$ where $M_{\theta}(\mathbf{x})\neq\mathbf{y}^+$
    \Ensure $\mathbf{x}^\prime$
    \State Initialize $\mathbf{z}^\prime \gets \mathbf{x}$ 
    \State Generate $\left\{\hat{\mathbf{x}}_{\theta,\mathbf{y}^+}\right\}_{n_{\mathcal{B}}} \gets p_{\theta}(\mathbf{x}_{\mathbf{y}^+})$ \Comment{Generate $n_{\mathcal{B}}$ samples using SGLD (Equation~\ref{eq:sgld}).}
    \State Store $\widehat{\mathbf{X}}_{\theta,\mathbf{y}^+} \gets \left\{\hat{\mathbf{x}}_{\theta,\mathbf{y}^+}\right\}_{n_{\mathcal{B}}}$ \Comment{Choose $n_E$ lowest-energy samples.}
    \State Run \textit{SCP} for $M_{\theta}$ using $\mathcal{D}$ \Comment{Calibrate model through split conformal prediction.}
    \State Initialize $t \gets 0$
    \While{\textit{not converged} or $t < T$} \Comment{For convergence conditions see below.}
    \State $\mathbf{z}^\prime \gets \mathbf{z}^\prime - \eta \nabla_{\mathbf{z}^\prime} \mathcal{L}(\mathbf{z}^\prime,\mathbf{y}^+,\widehat{\mathbf{X}}_{\theta,\mathbf{y}^+}; \Lambda, \alpha)$ \Comment{Take gradient step of size $\eta$.}
    \State $t \gets t+1$
    \EndWhile
    \State $\mathbf{x}^\prime \gets \mathbf{z}^\prime$
  \end{algorithmic}
\end{algorithm*}

\subsubsection{A Note on Convergence}\label{convergence}

Convergence is not typically discussed much in the context of CE, even though it has important implications on outcomes. One intuitive way to specify convergence is in terms of threshold probabilities: once the predicted probability $p(\mathbf{y}^+|\mathbf{x}^{\prime})$ exceeds some user-defined threshold $\gamma$ such that the counterfactual is valid, we could consider the search to have converged. In the binary case, for example, convergence could be defined as $p(\mathbf{y}^+|\mathbf{x}^{\prime})>0.5$ in this sense. Note, however, how this can be expected to yield counterfactuals in the proximity of the decision boundary, a region characterized by high aleatoric uncertainty. In other words, counterfactuals generated in this way would generally not be plausible. To avoid this from happening, we specify convergence in terms of gradients approaching zero for all our experiments and all of our generators. This is allows us to get a cleaner read on how the different counterfactual search objectives affect counterfactual outcomes. 

\subsubsection{\texttt{ECCCo.jl}}

We have built a small Julia package, \texttt{ECCCo.jl}, that was used to perform the analysis in this paper: https://github.com/pat-alt/ECCCo.jl. The package will be absorbed by \texttt{CounterfactualExplanations.jl} in the future~\citep{altmeyer2023explaining}.

\subsection{Experimental Setup}\label{app:setup}

In our experiments, we always generate multiple counterfactuals for each model and generator over multiple runs to account for stochasticity. For each generator, model and dataset, we sample a total of $n_f=100$ factuals for a single benchmark, where each time the factual and target class is drawn randomly. This process is repeated over 50 runs for the synthetic and tabular datasets, which yields a total of around 1.5 million counterfactuals per dataset for the final results. For vision datasets, we relied on 10 runs yielding a total of 300,000 counterfactuals. The final results presented in the paper and this appendix show the sample averages and the standard deviations of samples averages across all runs. For grid search, we used single runs for each parameter specification generating 100 counterfactuals per generator and model for synthetic and tabular datasets. For vision data, we generated 10 counterfactuals for each model, generator and parameter specification.

Table~\ref{tab:params} provides an overview of all parameters related to our experiments. The \textit{GMSC} data were randomly undersampled for balancing purposes and all features were standardized. \textit{MNIST} data was also randomly undersampled for reasons outlined below. Pixel values were preprocessed to fall in the range of $[-1,1]$ and a small Gaussian noise component ($\sigma=0.03$) was added to training samples following common practice in the EBM literature. All of our models were trained through mini-batch training using the Adam optimiser (\citet{kingma2014adam}). Table~\ref{tab:perf} shows standard evaluation metrics measuring the predictive performance of our different models grouped by dataset. These measures were computed on test data. 

Table~\ref{tab:genparams} summarises our hyperparameter choices for the counterfactual generators where $\eta$ denotes the learning rate used for Stochastic Gradient Descent (SGD) and $\lambda_1$, $\lambda_2$, $\lambda_3$ represent the chosen penalty strengths (Equations~\ref{eq:general} and~\ref{eq:eccco}). Here $\lambda_1$ also refers to the chosen penalty for the distance from factual values that applies to both \textit{Wachter} and \textit{REVISE}, but not \textit{Schut} which is penalty-free. \textit{Schut} is also the only generator that uses JSMA instead of SGD for optimization.

\import{../contents/}{table_params.tex}

\import{../contents/}{table_perf.tex}

\import{../contents/}{table_gen_params.tex}

\subsection{Compute}

Research reported in this work was partially or completely facilitated by computational resources and support of the DelftBlue~\citep{DHPC2022} and the Delft AI Cluster (DAIC: https://doc.daic.tudelft.nl/) at TU Delft. 

For grid search, we used 300 CPUs for tabular real-world datasets ($<$1.5 hours each), 150 CPUs for synthetic datasets ($\approx$1 hour each) and 100 CPUs for vision datasets ($<$4 hours each), where we ran a smaller grid search for the latter. To generate the final results reported in the tables we used 300 CPUs for all datasets ($<$1.5 hours each) except the vision datasets. For the latter, we used 50 CPUs for smaller final experiments and at longer run times ($\approx$5 hours each).

\subsection{Results}\label{app:results}

\begin{figure}
  \centering
  \includegraphics[width=1.0\linewidth]{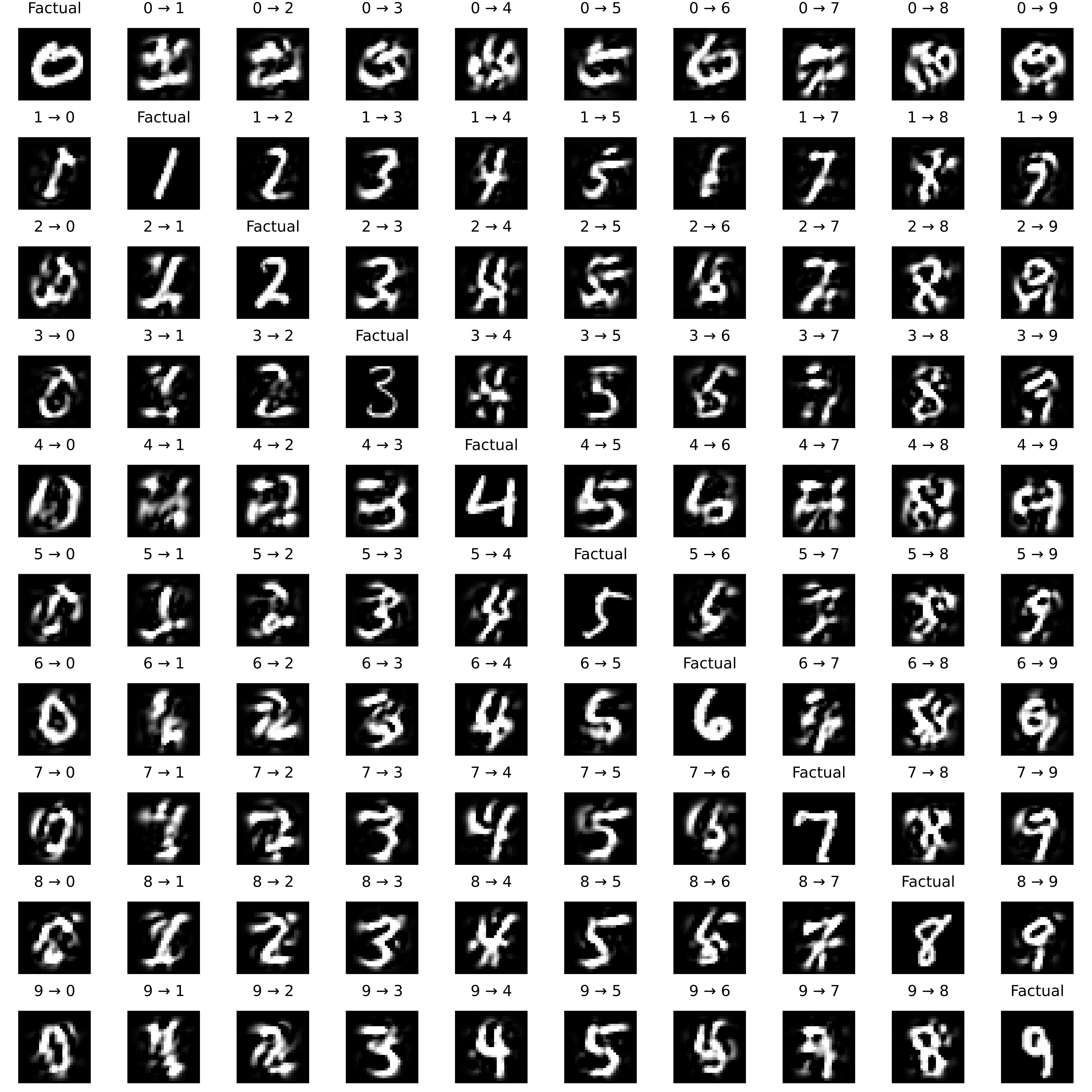}
  \caption{Counterfactuals for \textit{MNIST} data generated by \textit{ECCCo+}. The underlying model is a LeNet-5 \textit{CNN}. Original images are shown on the diagonal with the corresponding counterfactuals plotted across rows.}\label{fig:mnist-eccco-lenet}
\end{figure}

\begin{figure}
  \centering
  \includegraphics[width=1.0\linewidth]{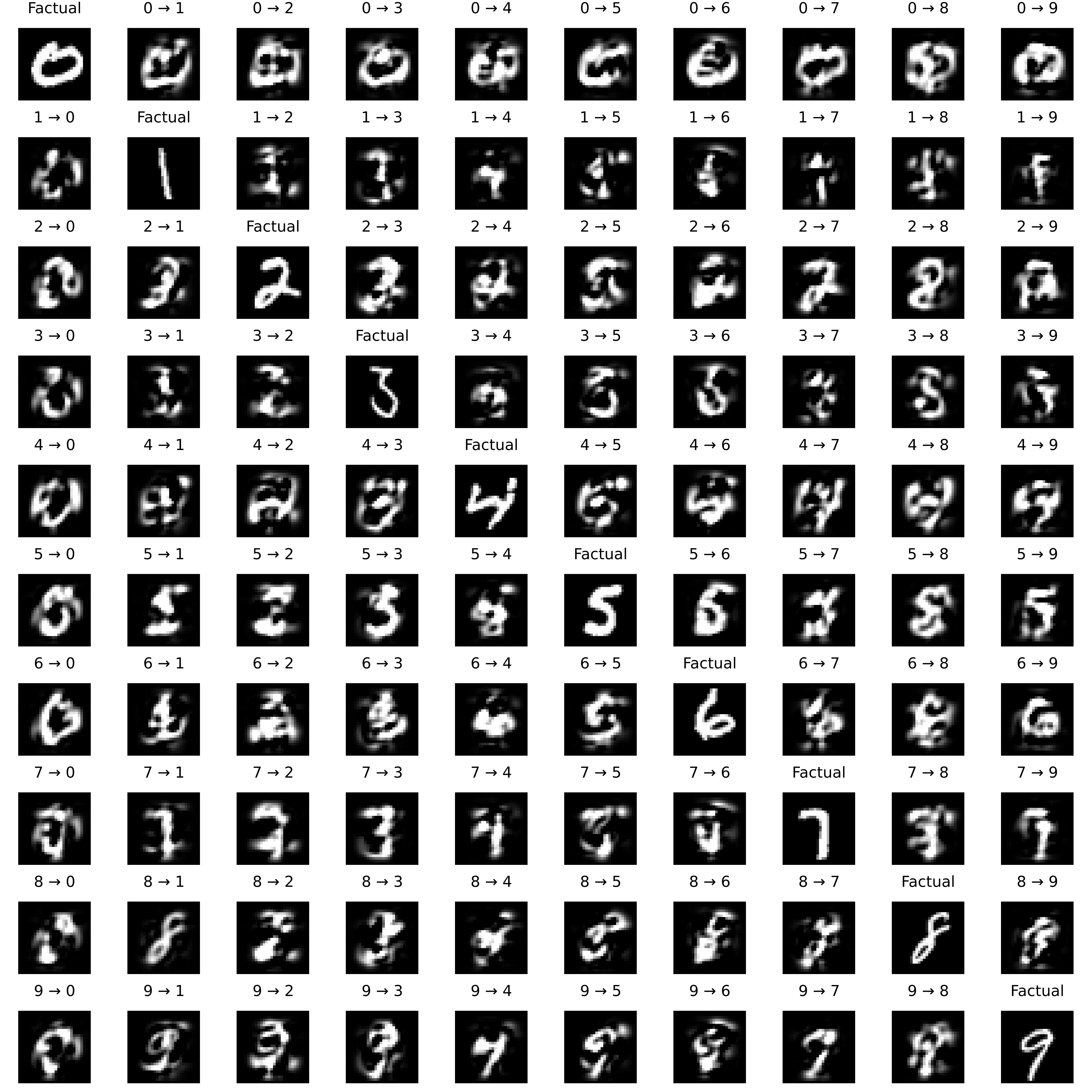}
  \caption{Counterfactuals for \textit{MNIST} data generated by \textit{ECCCo+}. The underlying model is an \textit{MLP}. Original images are shown on the diagonal with the corresponding counterfactuals plotted across rows.}\label{fig:mnist-eccco-mlp}
\end{figure}

\begin{figure}
  \centering
  \includegraphics[width=1.0\linewidth]{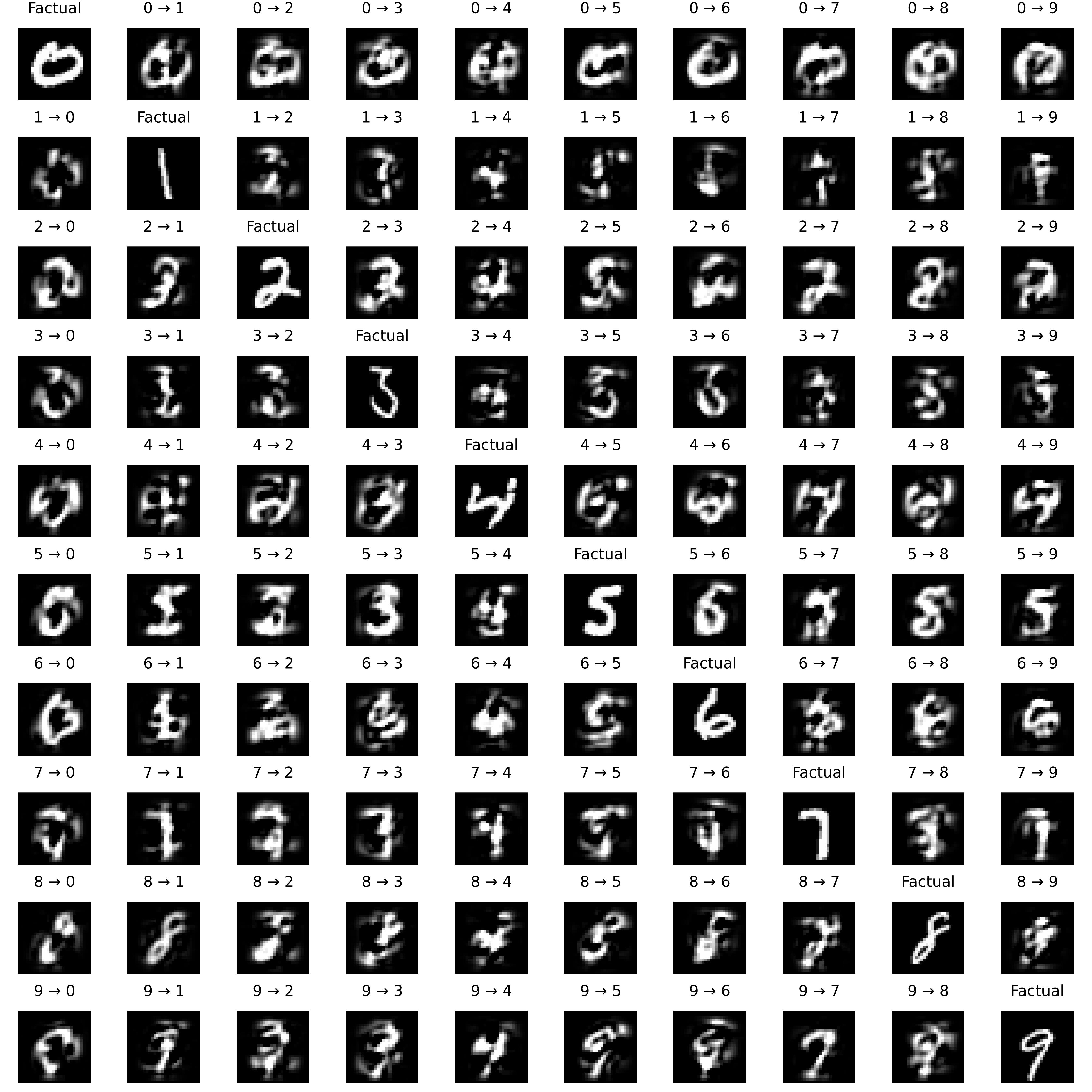}
  \caption{Counterfactuals for \textit{MNIST} data generated by \textit{ECCCo+}. The underlying model is an \textit{MLP} ensemble. Original images are shown on the diagonal with the corresponding counterfactuals plotted across rows.}\label{fig:mnist-eccco-mlp-ens}
\end{figure}

\begin{figure}
  \centering
  \includegraphics[width=1.0\linewidth]{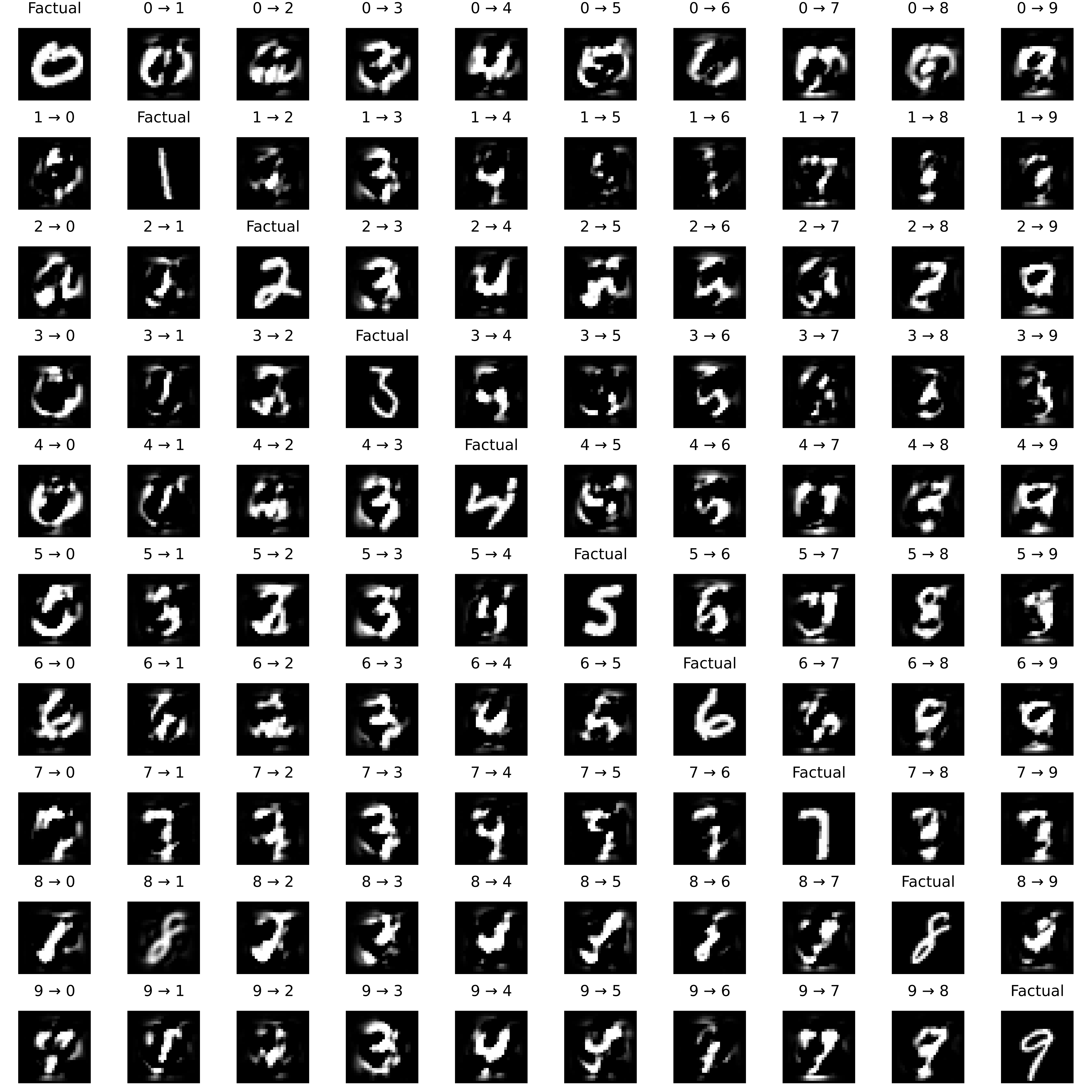}
  \caption{Counterfactuals for \textit{MNIST} data generated by \textit{ECCCo+}. The underlying model is a \textit{JEM}. Original images are shown on the diagonal with the corresponding counterfactuals plotted across rows.}\label{fig:mnist-eccco-jem}
\end{figure}

\begin{figure}
  \centering
  \includegraphics[width=1.0\linewidth]{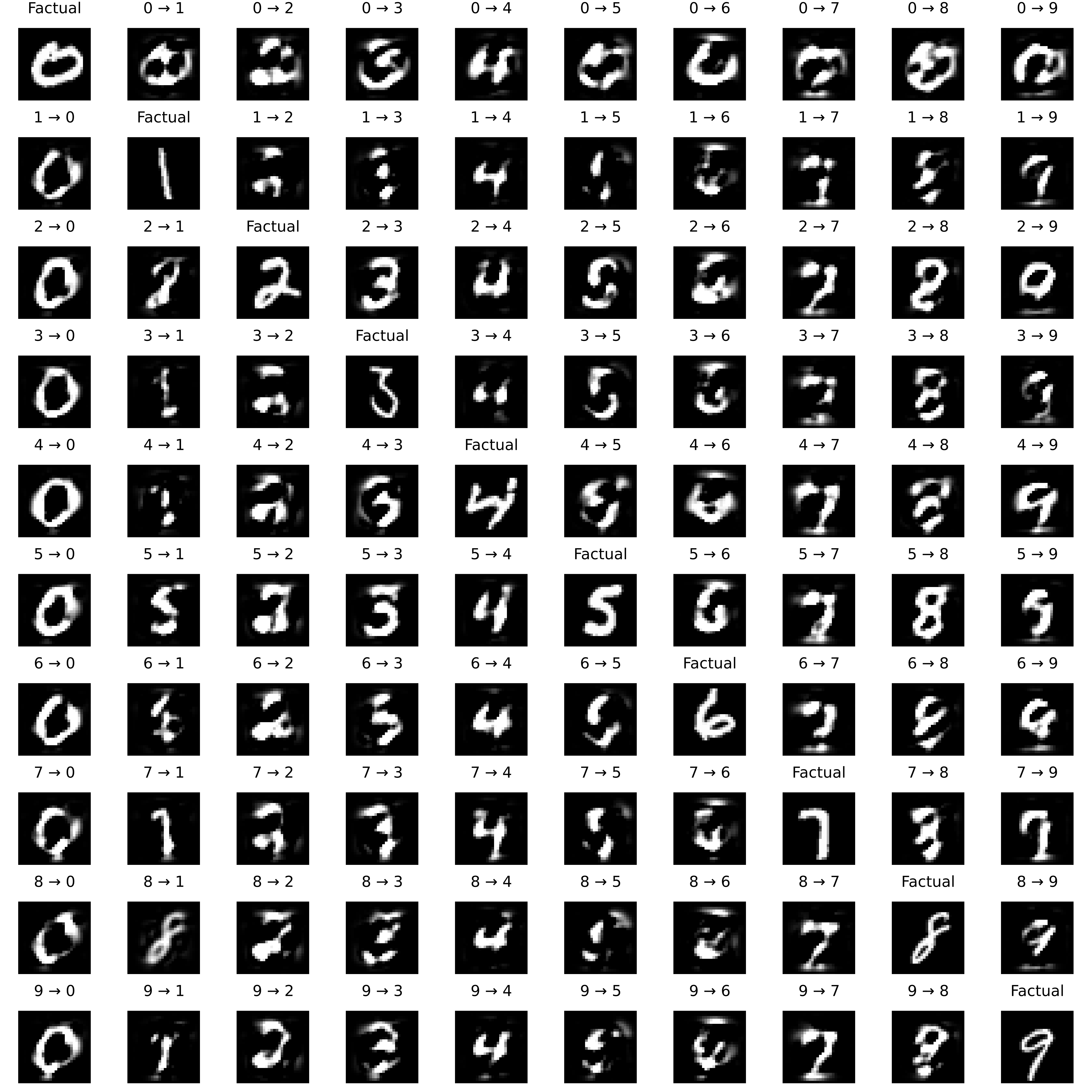}
  \caption{Counterfactuals for \textit{MNIST} data generated by \textit{ECCCo+}. The underlying model is a \textit{JEM} ensemble. Original images are shown on the diagonal with the corresponding counterfactuals plotted across rows.}\label{fig:mnist-eccco-jem-ens}
\end{figure}

\begin{figure}
  \centering
  \includegraphics[width=1.0\linewidth]{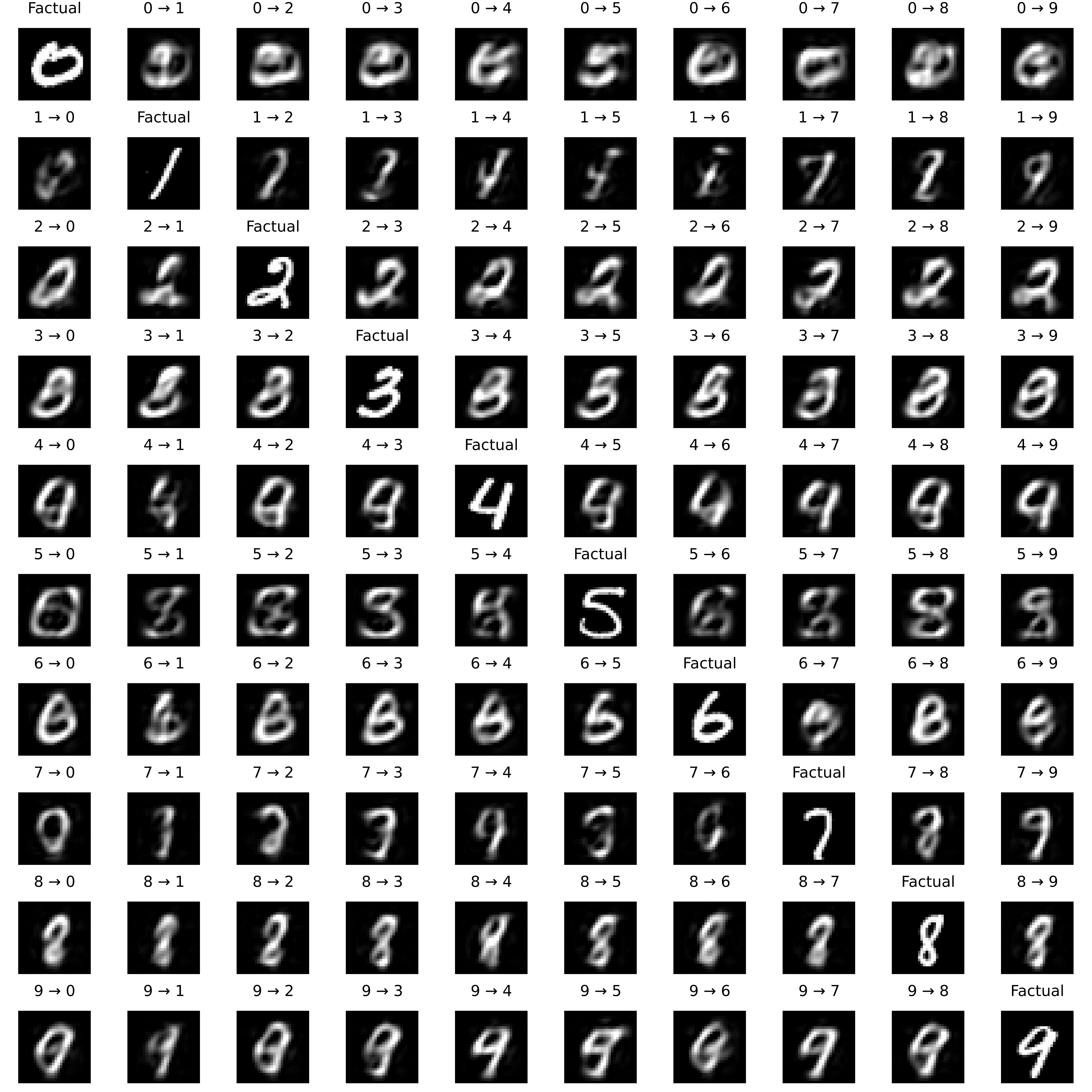}
  \caption{Counterfactuals for \textit{MNIST} data generated by \textit{REVISE}. The underlying model is a LeNet-5 \textit{CNN}. Original images are shown on the diagonal with the corresponding counterfactuals plotted across rows.}\label{fig:mnist-revise-lenet}
\end{figure}

\begin{figure}
  \centering
  \includegraphics[width=1.0\linewidth]{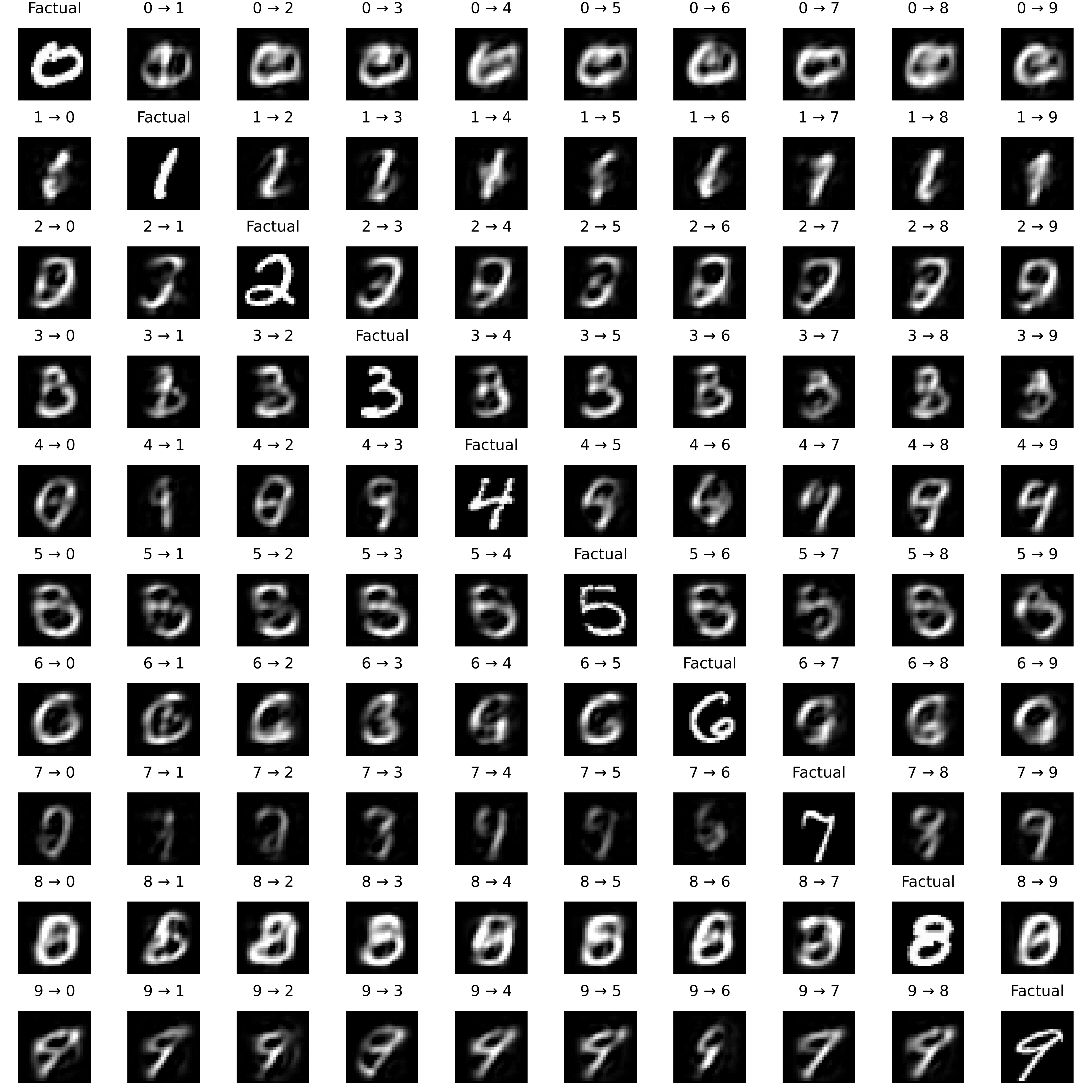}
  \caption{Counterfactuals for \textit{MNIST} data generated by \textit{REVISE}. The underlying model is an \textit{MLP}. Original images are shown on the diagonal with the corresponding counterfactuals plotted across rows.}\label{fig:mnist-revise-mlp}
\end{figure}

\begin{figure}
  \centering
  \includegraphics[width=1.0\linewidth]{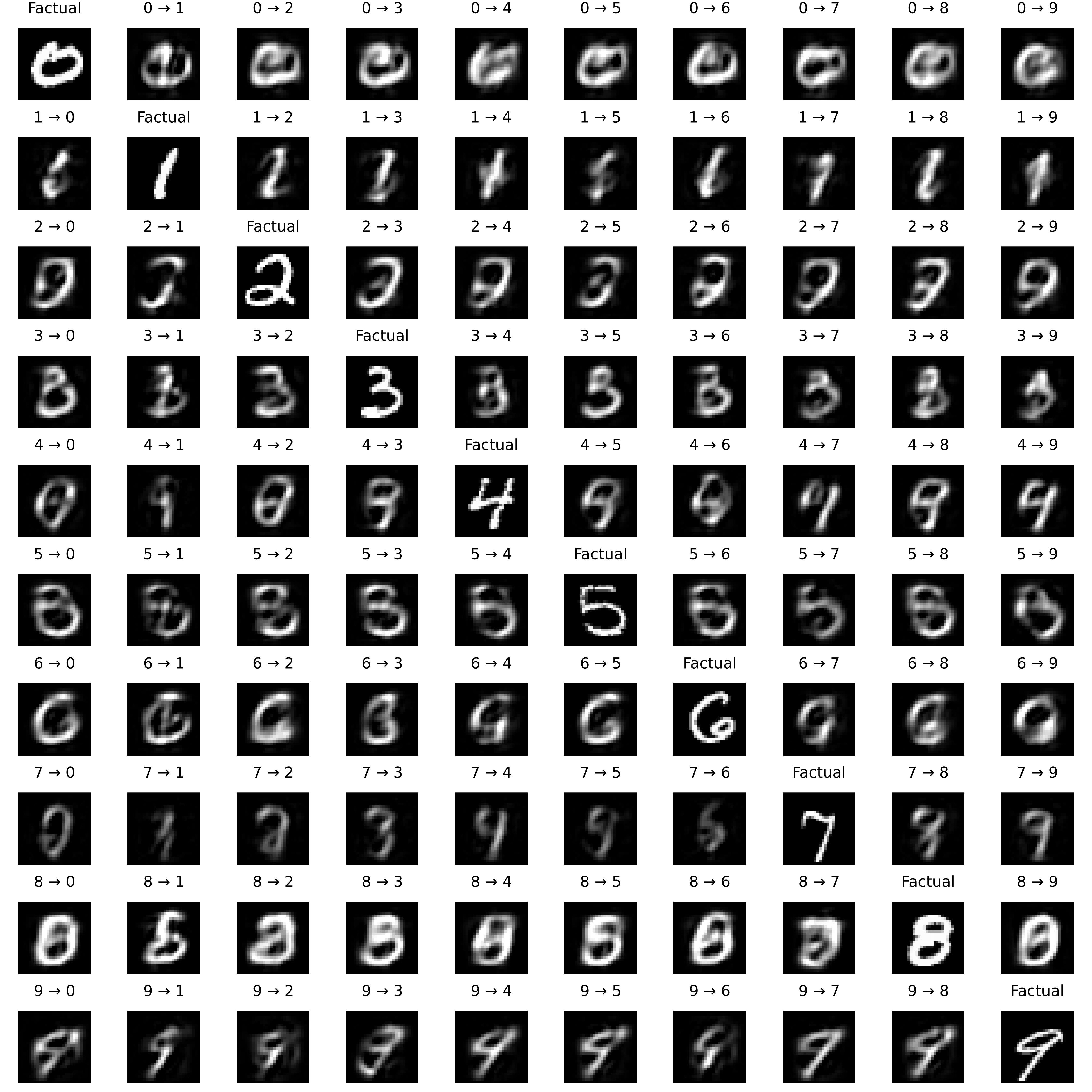}
  \caption{Counterfactuals for \textit{MNIST} data generated by \textit{REVISE}. The underlying model is an \textit{MLP} ensemble. Original images are shown on the diagonal with the corresponding counterfactuals plotted across rows.}\label{fig:mnist-revise-mlp-ens}
\end{figure}

\begin{figure}
  \centering
  \includegraphics[width=1.0\linewidth]{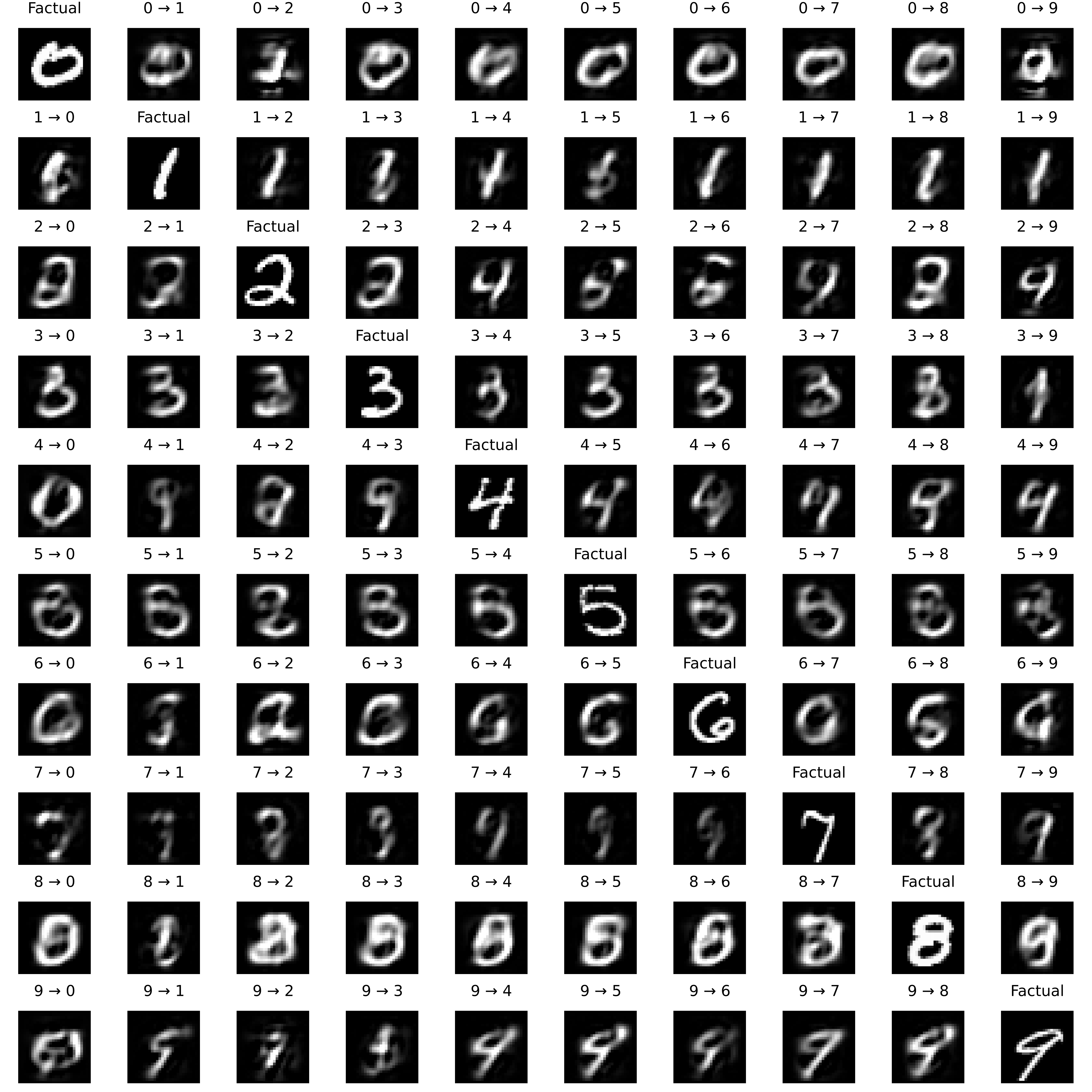}
  \caption{Counterfactuals for \textit{MNIST} data generated by \textit{REVISE}. The underlying model is a \textit{JEM}. Original images are shown on the diagonal with the corresponding counterfactuals plotted across rows.}\label{fig:mnist-revise-jem}
\end{figure}

\begin{figure}
  \centering
  \includegraphics[width=1.0\linewidth]{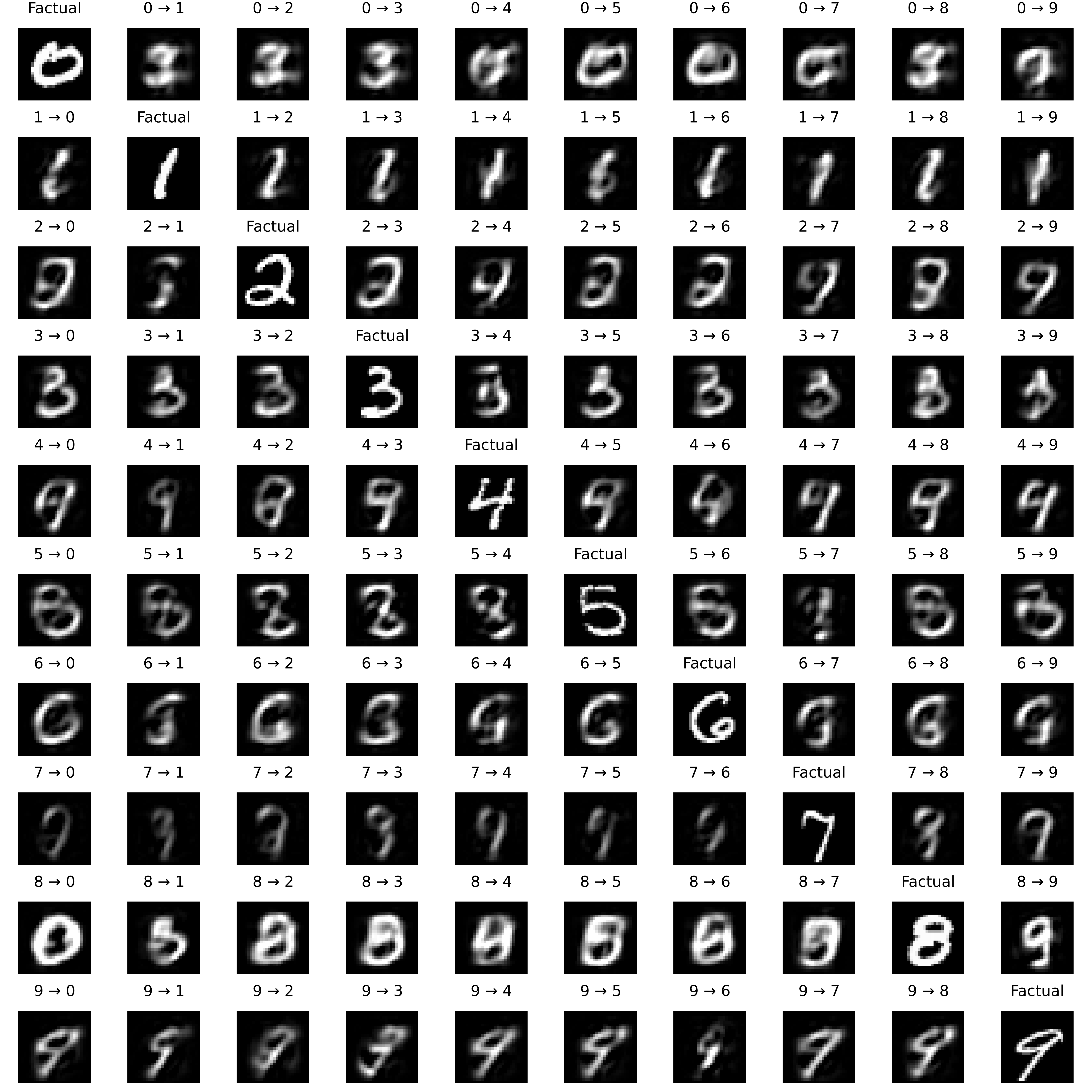}
  \caption{Counterfactuals for \textit{MNIST} data generated by \textit{REVISE}. The underlying model is a \textit{JEM} ensemble. Original images are shown on the diagonal with the corresponding counterfactuals plotted across rows.}\label{fig:mnist-revise-jem-ens}
\end{figure}

\begin{figure}[h]
  \centering
  \includegraphics[width=1.0\linewidth]{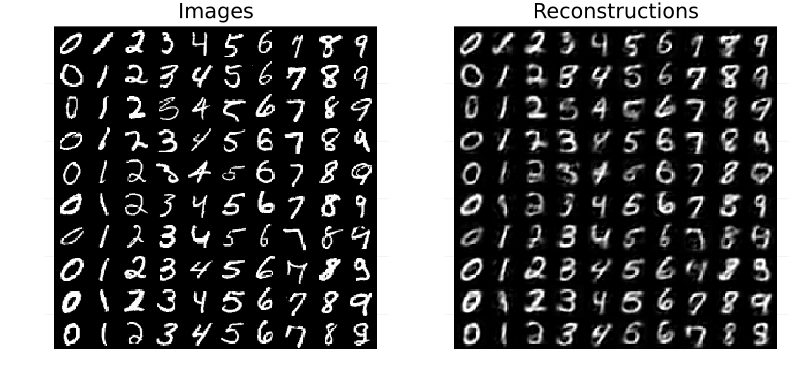}
  \caption{Randomly drawn \textit{MNIST} images and their reconstructions generated by the VAE used by \textit{REVISE}.}\label{fig:vae-rec}
\end{figure}

Figures~\ref{fig:mnist-eccco-lenet} to~\ref{fig:mnist-eccco-jem-ens} show examples of counterfactuals for \textit{MNIST} generated by \textit{ECCCo+} for our different models. Original images are shown on the diagonal and the corresponding counterfactuals are plotted across rows. Figures~\ref{fig:mnist-revise-lenet} to~\ref{fig:mnist-revise-jem-ens} show the same examples but for \textit{REVISE}. Both counterfactual generators have access to the same optimizer. While the results for \textit{REVISE} look fairly poor here, we have observed better results for optizers with higher step sizes. Note that the seemingly poor performance by \textit{REVISE} upon visual inspection is not driven by a weak surrogate VAE: Figure~\ref{fig:vae-rec} shows image reconstructions generated by the VAE.


Tables~\ref{tab:results-linearly-separable} to~\ref{tab:results-fashion-mnist} reports all of the evaluation metrics we have computed. Tables~\ref{tab:results-linearly-separable-valid} to~\ref{tab:results-fashion-mnist-valid} reports the same metrics for the subset of valid counterfactuals. The `Unfaithfulness' and `Implausibility' metrics have been discussed extensively in the body of the paper. The `Cost' metric relates to the distance between the factual and the counterfactual and is measured using the L1 Norm. The `Redundancy' metric measures sparsity in is defined as the percentage of features that remain unperturbed (higher is better). The `Uncertainty' metric is just the average value of the smooth set size penalty (Equation~\ref{eq:setsize}). Finally, `Validity' is the percentage of valid counterfactuals. 

\import{../contents/}{table-linearly-separable.tex}

\import{../contents/}{table-circles.tex}

\import{../contents/}{table-moons.tex}

\import{../contents/}{table-california-housing.tex}

\import{../contents/}{table-gmsc.tex}

\import{../contents/}{table-german-credit.tex}

\import{../contents/}{table-mnist.tex}

\import{../contents/}{table-fashion-mnist.tex}

\import{../contents/}{table-linearly-separable-valid.tex}

\import{../contents/}{table-circles-valid.tex}

\import{../contents/}{table-moons-valid.tex}

\import{../contents/}{table-california-housing-valid.tex}

\import{../contents/}{table-gmsc-valid.tex}

\import{../contents/}{table-german-credit-valid.tex}

\import{../contents/}{table-mnist-valid.tex}

\import{../contents/}{table-fashion-mnist-valid.tex}

%% file: contents/table_ebm_params.tex
\begin{table}

\caption{EBM hyperparemeter choices for our experiments. \label{tab:ebmparams} \newline}
\centering
\fontsize{8}{10}\selectfont
\begin{tabular}[t]{rrrr}
\toprule
Dataset & SGLD Steps & Batch Size & $\lambda$\\
\midrule
Linearly Separable & 50 & 50 & 0.10\\
Moons & 30 & 10 & 0.10\\
Circles & 30 & 50 & 0.01\\
California Housing & 30 & 10 & 0.10\\
GMSC & 30 & 10 & 0.10\\
German Credit & 30 & 10 & 0.10\\
MNIST & 25 & 10 & 0.01\\
Fashion MNIST & 25 & 10 & 0.01\\
\bottomrule
\end{tabular}
\end{table}

%% file: contents/table_params.tex
\begin{table}

\caption{Paremeter choices for our experiments. \label{tab:params} \newline}
\centering
\resizebox{\linewidth}{!}{
\begin{tabular}[t]{rrrrrrrr}
\toprule
\multicolumn{2}{c}{ } & \multicolumn{4}{c}{Network Architecture} & \multicolumn{2}{c}{Training} \\
\cmidrule(l{3pt}r{3pt}){3-6} \cmidrule(l{3pt}r{3pt}){7-8}
Dataset & Sample Size & Hidden Units & Hidden Layers & Activation & Ensemble Size & Epochs & Batch Size\\
\midrule
Linearly Separable & 1000 & 16 & 3 & swish & 5 & 100 & 100\\
Moons & 2500 & 32 & 3 & relu & 5 & 500 & 128\\
Circles & 1000 & 32 & 3 & swish & 5 & 100 & 100\\
California Housing & 16500 & 32 & 3 & relu & 5 & 100 & 128\\
GMSC & 13370 & 32 & 3 & relu & 5 & 100 & 128\\
German Credit & 800 & 32 & 3 & relu & 5 & 100 & 80\\
MNIST & 10000 & 32 & 1 & relu & 5 & 100 & 128\\
Fashion MNIST & 10000 & 32 & 2 & relu & 5 & 100 & 128\\
\bottomrule
\end{tabular}}
\end{table}

%% file: contents/table_perf.tex
\begin{table}

\caption{Various standard performance metrics for our different models grouped by dataset. \label{tab:perf} \newline}
\centering
\fontsize{8}{10}\selectfont
\begin{tabular}[t]{rrrrr}
\toprule
\multicolumn{2}{c}{ } & \multicolumn{3}{c}{Performance Metrics} \\
\cmidrule(l{3pt}r{3pt}){3-5}
Dataset & Model & Accuracy & Precision & F1-Score\\
\midrule
 & JEM & 0.98 & 0.98 & 0.98\\

 & JEM Ensemble & 0.99 & 0.99 & 0.99\\

 & MLP & 0.99 & 0.99 & 0.99\\

\multirow[t]{-4}{*}{\raggedleft\arraybackslash Linearly Separable} & MLP Ensemble & 0.99 & 0.99 & 0.99\\
\cmidrule{1-5}
 & JEM & 1.00 & 1.00 & 1.00\\

 & JEM Ensemble & 1.00 & 1.00 & 1.00\\

 & MLP & 1.00 & 1.00 & 1.00\\

\multirow[t]{-4}{*}{\raggedleft\arraybackslash Moons} & MLP Ensemble & 1.00 & 1.00 & 1.00\\
\cmidrule{1-5}
 & JEM & 1.00 & 1.00 & 1.00\\

 & JEM Ensemble & 1.00 & 1.00 & 1.00\\

 & MLP & 1.00 & 1.00 & 1.00\\

\multirow[t]{-4}{*}{\raggedleft\arraybackslash Circles} & MLP Ensemble & 1.00 & 1.00 & 1.00\\
\cmidrule{1-5}
 & JEM & 0.86 & 0.86 & 0.86\\

 & JEM Ensemble & 0.86 & 0.86 & 0.86\\

 & MLP & 0.88 & 0.88 & 0.88\\

\multirow[t]{-4}{*}{\raggedleft\arraybackslash California Housing} & MLP Ensemble & 0.88 & 0.88 & 0.88\\
\cmidrule{1-5}
 & JEM & 0.73 & 0.75 & 0.73\\

 & JEM Ensemble & 0.74 & 0.75 & 0.74\\

 & MLP & 0.75 & 0.75 & 0.75\\

\multirow[t]{-4}{*}{\raggedleft\arraybackslash GMSC} & MLP Ensemble & 0.75 & 0.75 & 0.75\\
\cmidrule{1-5}
 & JEM & 0.58 & 0.64 & 0.52\\

 & JEM Ensemble & 0.54 & 0.72 & 0.43\\

 & MLP & 0.52 & 0.75 & 0.37\\

\multirow[t]{-4}{*}{\raggedleft\arraybackslash German Credit} & MLP Ensemble & 0.51 & 0.75 & 0.36\\
\cmidrule{1-5}
 & JEM & 0.84 & 0.85 & 0.84\\

 & JEM Ensemble & 0.90 & 0.90 & 0.90\\

 & LeNet-5 & 0.98 & 0.98 & 0.98\\

 & MLP & 0.95 & 0.95 & 0.95\\

\multirow[t]{-5}{*}{\raggedleft\arraybackslash MNIST} & MLP Ensemble & 0.95 & 0.95 & 0.95\\
\cmidrule{1-5}
 & JEM & 0.62 & 0.70 & 0.62\\

 & JEM Ensemble & 0.78 & 0.78 & 0.78\\

 & LeNet-5 & 0.83 & 0.84 & 0.82\\

 & MLP & 0.82 & 0.83 & 0.82\\

\multirow[t]{-5}{*}{\raggedleft\arraybackslash Fashion MNIST} & MLP Ensemble & 0.84 & 0.84 & 0.84\\
\bottomrule
\end{tabular}
\end{table}

%% file: contents/table_gen_params.tex
\begin{table}

\caption{Generator hyperparameters: the optimiser step size ($\eta$); penalty strengths where $\lambda_1$ applies to all generators but \textit{Schut} and the other parameter are specific to \textit{ECCCo}; finally, the strength for the Ridge penalty on energy for \textit{ECCCo}.\label{tab:genparams} \newline}
\centering
\fontsize{8}{10}\selectfont
\begin{tabular}[t]{rrrrrr}
\toprule
Dataset & $\eta$ & $\lambda_1$ & $\lambda_2$ & $\lambda_3$ & Ridge penalty\\
\midrule
Linearly Separable & 0.05 & 0.10 & 0.1 & 0.5 & 0.0\\
Moons & 0.05 & 0.10 & 0.1 & 0.2 & 0.0\\
Circles & 0.05 & 0.10 & 0.1 & 0.2 & 0.0\\
California Housing & 0.05 & 0.10 & 0.1 & 0.5 & 0.0\\
GMSC & 0.05 & 0.10 & 0.1 & 0.5 & 0.0\\
German Credit & 0.05 & 0.10 & 0.1 & 0.1 & 0.5\\
MNIST & 0.05 & 0.01 & 0.1 & 0.3 & 0.0\\
Fashion MNIST & 0.05 & 0.01 & 0.1 & 0.3 & 0.0\\
\bottomrule
\end{tabular}
\end{table}

%% file: contents/table-linearly-separable.tex
\begin{table}

\caption{All results for Linearly Separable dataset: sample averages +/- one standard deviation over all counterfactuals. Best outcomes are highlighted in bold. Asterisks indicate that the given value is more than one (*) or two (**) standard deviations away from the baseline (\textit{Wachter}). \label{tab:results-linearly-separable} \newline}
\centering
\resizebox{\linewidth}{!}{
\begin{tabular}[t]{cccccccc}
\toprule
Model & Generator & Unfaithfulness ↓ & Implausibility ↓ & Uncertainty ↓ & Cost ↓ & Redundancy ↑ & Validity ↑\\
\midrule
 & ECCCo-L1 & 0.06 ± 0.01** & \textbf{0.16 ± 0.02}** & 0.00 ± 0.00\hphantom{*}\hphantom{*} & 0.93 ± 0.02\hphantom{*}\hphantom{*} & 0.00 ± 0.00\hphantom{*}\hphantom{*} & \textbf{1.00 ± 0.00}\hphantom{*}\hphantom{*}\\

 & ECCCo-L1 (no CP) & \textbf{0.06 ± 0.01}** & 0.16 ± 0.02** & 0.00 ± 0.00\hphantom{*}\hphantom{*} & 0.93 ± 0.02\hphantom{*}\hphantom{*} & 0.00 ± 0.00\hphantom{*}\hphantom{*} & \textbf{1.00 ± 0.00}\hphantom{*}\hphantom{*}\\

 & ECCCo-L1 (no EBM) & 0.09 ± 0.01** & 0.24 ± 0.01\hphantom{*}\hphantom{*} & 0.00 ± 0.00\hphantom{*}\hphantom{*} & 0.88 ± 0.01\hphantom{*}\hphantom{*} & 0.00 ± 0.00\hphantom{*}\hphantom{*} & \textbf{1.00 ± 0.00}\hphantom{*}\hphantom{*}\\

 & ECCCo & 0.19 ± 0.01** & 0.19 ± 0.01** & 0.00 ± 0.00\hphantom{*}\hphantom{*} & 0.85 ± 0.01** & \textbf{0.00 ± 0.00}\hphantom{*}\hphantom{*} & \textbf{1.00 ± 0.00}\hphantom{*}\hphantom{*}\\

 & ECCCo (no CP) & 0.19 ± 0.01** & 0.19 ± 0.01** & 0.00 ± 0.00\hphantom{*}\hphantom{*} & 0.85 ± 0.01** & \textbf{0.00 ± 0.00}\hphantom{*}\hphantom{*} & \textbf{1.00 ± 0.00}\hphantom{*}\hphantom{*}\\

 & ECCCo (no EBM) & 0.24 ± 0.01\hphantom{*}\hphantom{*} & 0.24 ± 0.01\hphantom{*}\hphantom{*} & 0.00 ± 0.00\hphantom{*}\hphantom{*} & 0.88 ± 0.01\hphantom{*}\hphantom{*} & 0.00 ± 0.00\hphantom{*}\hphantom{*} & \textbf{1.00 ± 0.00}\hphantom{*}\hphantom{*}\\

 & REVISE & 0.27 ± 0.01\hphantom{*}\hphantom{*} & 0.18 ± 0.00** & 0.00 ± 0.00\hphantom{*}\hphantom{*} & \textbf{0.42 ± 0.01}** & 0.00 ± 0.00\hphantom{*}\hphantom{*} & 0.47 ± 0.00\hphantom{*}\hphantom{*}\\
\bottomrule
\end{tabular}}
\end{table}

%% file: contents/table-circles.tex
\begin{table}

\caption{All results for Circles dataset: sample averages +/- one standard deviation over all counterfactuals. Best outcomes are highlighted in bold. Asterisks indicate that the given value is more than one (*) or two (**) standard deviations away from the baseline (\textit{Wachter}). \label{tab:results-circles} \newline}
\centering
\resizebox{\linewidth}{!}{
\begin{tabular}[t]{cccccccc}
\toprule
Model & Generator & Unfaithfulness ↓ & Implausibility ↓ & Uncertainty ↓ & Cost ↓ & Redundancy ↑ & Validity ↑\\
\midrule
 & ECCCo-L1 & 0.28 ± 0.01** & 0.68 ± 0.01\hphantom{*}\hphantom{*} & 0.00 ± 0.00\hphantom{*}\hphantom{*} & 0.70 ± 0.01\hphantom{*}\hphantom{*} & 0.00 ± 0.00\hphantom{*}\hphantom{*} & 0.53 ± 0.01\hphantom{*}\hphantom{*}\\

 & ECCCo-L1 (no CP) & \textbf{0.27 ± 0.01}** & 0.68 ± 0.01\hphantom{*}\hphantom{*} & 0.00 ± 0.00\hphantom{*}\hphantom{*} & 0.70 ± 0.01\hphantom{*}\hphantom{*} & 0.00 ± 0.00\hphantom{*}\hphantom{*} & 0.53 ± 0.01\hphantom{*}\hphantom{*}\\

 & ECCCo-L1 (no EBM) & 0.28 ± 0.01** & 0.68 ± 0.01\hphantom{*}\hphantom{*} & 0.00 ± 0.00\hphantom{*}\hphantom{*} & 0.70 ± 0.01\hphantom{*}\hphantom{*} & 0.00 ± 0.00\hphantom{*}\hphantom{*} & 0.53 ± 0.01\hphantom{*}\hphantom{*}\\

 & ECCCo & 0.69 ± 0.03\hphantom{*}\hphantom{*} & 0.69 ± 0.00\hphantom{*}\hphantom{*} & 0.00 ± 0.00\hphantom{*}\hphantom{*} & 0.76 ± 0.02\hphantom{*}\hphantom{*} & 0.00 ± 0.00\hphantom{*}\hphantom{*} & 0.53 ± 0.00\hphantom{*}\hphantom{*}\\

 & ECCCo (no CP) & 0.68 ± 0.03\hphantom{*}\hphantom{*} & 0.69 ± 0.00\hphantom{*}\hphantom{*} & 0.00 ± 0.00\hphantom{*}\hphantom{*} & 0.76 ± 0.02\hphantom{*}\hphantom{*} & 0.00 ± 0.00\hphantom{*}\hphantom{*} & 0.53 ± 0.00\hphantom{*}\hphantom{*}\\

 & ECCCo (no EBM) & 0.69 ± 0.02\hphantom{*}\hphantom{*} & 0.68 ± 0.01\hphantom{*}\hphantom{*} & 0.00 ± 0.00\hphantom{*}\hphantom{*} & 0.70 ± 0.01\hphantom{*}\hphantom{*} & 0.00 ± 0.00\hphantom{*}\hphantom{*} & 0.53 ± 0.01\hphantom{*}\hphantom{*}\\

 & REVISE & 0.32 ± 0.01** & 0.68 ± 0.00\hphantom{*}\hphantom{*} & 0.00 ± 0.00\hphantom{*}\hphantom{*} & 0.94 ± 0.01\hphantom{*}\hphantom{*} & 0.00 ± 0.00\hphantom{*}\hphantom{*} & 0.47 ± 0.00\hphantom{*}\hphantom{*}\\
\bottomrule
\end{tabular}}
\end{table}

%% file: contents/table-moons.tex
\begin{table}

\caption{All results for Moons dataset: sample averages +/- one standard deviation over all counterfactuals. Best outcomes are highlighted in bold. Asterisks indicate that the given value is more than one (*) or two (**) standard deviations away from the baseline (\textit{Wachter}). \label{tab:results-moons} \newline}
\centering
\resizebox{\linewidth}{!}{
\begin{tabular}[t]{cccccccc}
\toprule
Model & Generator & Unfaithfulness ↓ & Implausibility ↓ & Uncertainty ↓ & Cost ↓ & Redundancy ↑ & Validity ↑\\
\midrule
 & ECCCo-L1 & 0.29 ± 0.02** & 0.42 ± 0.04\hphantom{*}\hphantom{*} & 0.14 ± 0.03\hphantom{*}\hphantom{*} & 1.02 ± 0.03\hphantom{*}\hphantom{*} & 0.00 ± 0.00\hphantom{*}\hphantom{*} & 1.00 ± 0.00\hphantom{*}\hphantom{*}\\

 & ECCCo-L1 (no CP) & \textbf{0.29 ± 0.02}** & 0.42 ± 0.04\hphantom{*}\hphantom{*} & 0.15 ± 0.03\hphantom{*}\hphantom{*} & \textbf{1.02 ± 0.03}\hphantom{*}\hphantom{*} & 0.00 ± 0.00\hphantom{*}\hphantom{*} & 1.00 ± 0.00\hphantom{*}\hphantom{*}\\

 & ECCCo-L1 (no EBM) & 0.30 ± 0.02** & 0.42 ± 0.04\hphantom{*}\hphantom{*} & 0.13 ± 0.03\hphantom{*}\hphantom{*} & 1.03 ± 0.02\hphantom{*}\hphantom{*} & 0.00 ± 0.00\hphantom{*}\hphantom{*} & 1.00 ± 0.00\hphantom{*}\hphantom{*}\\

 & ECCCo & 0.55 ± 0.06** & 0.15 ± 0.01** & \textbf{0.01 ± 0.01}** & 1.28 ± 0.04\hphantom{*}\hphantom{*} & 0.00 ± 0.00\hphantom{*}\hphantom{*} & \textbf{1.00 ± 0.00}**\\

 & ECCCo (no CP) & 0.57 ± 0.05** & \textbf{0.15 ± 0.01}** & 0.01 ± 0.01** & 1.28 ± 0.04\hphantom{*}\hphantom{*} & 0.00 ± 0.00\hphantom{*}\hphantom{*} & \textbf{1.00 ± 0.00}**\\

 & ECCCo (no EBM) & 0.69 ± 0.04\hphantom{*}\hphantom{*} & 0.42 ± 0.04\hphantom{*}\hphantom{*} & 0.13 ± 0.03\hphantom{*}\hphantom{*} & 1.03 ± 0.02\hphantom{*}\hphantom{*} & 0.00 ± 0.00\hphantom{*}\hphantom{*} & 1.00 ± 0.00\hphantom{*}\hphantom{*}\\

 & REVISE & 0.54 ± 0.05** & 0.31 ± 0.02** & 0.15 ± 0.06\hphantom{*}\hphantom{*} & 1.41 ± 0.12\hphantom{*}\hphantom{*} & 0.00 ± 0.00\hphantom{*}\hphantom{*} & 0.84 ± 0.07\hphantom{*}\hphantom{*}\\
\bottomrule
\end{tabular}}
\end{table}

%% file: contents/table-california-housing.tex
\begin{table}

\caption{All results for California Housing dataset: sample averages +/- one standard deviation over all counterfactuals. Best outcomes are highlighted in bold. Asterisks indicate that the given value is more than one (*) or two (**) standard deviations away from the baseline (\textit{Wachter}). \label{tab:results-california-housing} \newline}
\centering
\resizebox{\linewidth}{!}{
\begin{tabular}[t]{cccccccc}
\toprule
Model & Generator & Unfaithfulness ↓ & Implausibility ↓ & Uncertainty ↓ & Cost ↓ & Redundancy ↑ & Validity ↑\\
\midrule
 & ECCCo-L1 & 0.98 ± 0.05** & 1.04 ± 0.12\hphantom{*}\hphantom{*} & 0.24 ± 0.02\hphantom{*}\hphantom{*} & 1.22 ± 0.07\hphantom{*}\hphantom{*} & 0.00 ± 0.00\hphantom{*}\hphantom{*} & 0.98 ± 0.01\hphantom{*}\hphantom{*}\\

 & ECCCo-L1 (no CP) & \textbf{0.97 ± 0.05}** & 1.04 ± 0.12\hphantom{*}\hphantom{*} & 0.25 ± 0.02\hphantom{*}\hphantom{*} & 1.18 ± 0.07\hphantom{*}\hphantom{*} & 0.00 ± 0.00\hphantom{*}\hphantom{*} & 0.98 ± 0.01\hphantom{*}\hphantom{*}\\

 & ECCCo-L1 (no EBM) & 1.00 ± 0.06** & 1.07 ± 0.12\hphantom{*}\hphantom{*} & 0.24 ± 0.02\hphantom{*}\hphantom{*} & 1.12 ± 0.07\hphantom{*}\hphantom{*} & 0.01 ± 0.00\hphantom{*}\hphantom{*} & 0.98 ± 0.01\hphantom{*}\hphantom{*}\\

 & ECCCo & 1.94 ± 0.13*\hphantom{*} & 0.78 ± 0.08** & \textbf{0.18 ± 0.00}** & 2.05 ± 0.15\hphantom{*}\hphantom{*} & 0.01 ± 0.00\hphantom{*}\hphantom{*} & \textbf{1.00 ± 0.00}**\\

 & ECCCo+ & 1.84 ± 0.12** & 0.71 ± 0.07** & 0.19 ± 0.00** & 3.29 ± 0.18\hphantom{*}\hphantom{*} & 0.00 ± 0.00\hphantom{*}\hphantom{*} & 1.00 ± 0.01**\\

 & ECCCo (no CP) & 1.95 ± 0.14*\hphantom{*} & 0.78 ± 0.08** & 0.18 ± 0.00** & 2.03 ± 0.16\hphantom{*}\hphantom{*} & 0.00 ± 0.00\hphantom{*}\hphantom{*} & \textbf{1.00 ± 0.00}**\\

 & ECCCo (no EBM) & 2.13 ± 0.11\hphantom{*}\hphantom{*} & 1.07 ± 0.12\hphantom{*}\hphantom{*} & 0.24 ± 0.02\hphantom{*}\hphantom{*} & 1.12 ± 0.07\hphantom{*}\hphantom{*} & 0.01 ± 0.00\hphantom{*}\hphantom{*} & 0.98 ± 0.01\hphantom{*}\hphantom{*}\\
\bottomrule
\end{tabular}}
\end{table}

%% file: contents/table-gmsc.tex
\begin{table}

\caption{All results for GMSC dataset: sample averages +/- one standard deviation over all counterfactuals. Best outcomes are highlighted in bold. Asterisks indicate that the given value is more than one (*) or two (**) standard deviations away from the baseline (\textit{Wachter}). \label{tab:results-gmsc} \newline}
\centering
\resizebox{\linewidth}{!}{
\begin{tabular}[t]{cccccccc}
\toprule
Model & Generator & Unfaithfulness ↓ & Implausibility ↓ & Uncertainty ↓ & Cost ↓ & Redundancy ↑ & Validity ↑\\
\midrule
 & ECCCo-L1 & \textbf{0.94 ± 0.05}** & 1.11 ± 0.18\hphantom{*}\hphantom{*} & 0.29 ± 0.01** & 1.03 ± 0.04\hphantom{*}\hphantom{*} & 0.13 ± 0.01\hphantom{*}\hphantom{*} & 0.98 ± 0.01\hphantom{*}\hphantom{*}\\

 & ECCCo-L1 (no CP) & 0.96 ± 0.05** & 1.10 ± 0.18\hphantom{*}\hphantom{*} & 0.31 ± 0.01*\hphantom{*} & 0.96 ± 0.04\hphantom{*}\hphantom{*} & 0.14 ± 0.01\hphantom{*}\hphantom{*} & 0.98 ± 0.01\hphantom{*}\hphantom{*}\\

 & ECCCo-L1 (no EBM) & 0.97 ± 0.05** & 1.09 ± 0.18\hphantom{*}\hphantom{*} & 0.31 ± 0.01*\hphantom{*} & 0.89 ± 0.03\hphantom{*}\hphantom{*} & 0.14 ± 0.01\hphantom{*}\hphantom{*} & 0.98 ± 0.01\hphantom{*}\hphantom{*}\\

 & ECCCo & 1.85 ± 0.12\hphantom{*}\hphantom{*} & 1.01 ± 0.18\hphantom{*}\hphantom{*} & 0.34 ± 0.01\hphantom{*}\hphantom{*} & 1.12 ± 0.05\hphantom{*}\hphantom{*} & 0.13 ± 0.01\hphantom{*}\hphantom{*} & \textbf{0.99 ± 0.01}*\hphantom{*}\\

 & ECCCo+ & 1.72 ± 0.13*\hphantom{*} & 0.90 ± 0.18*\hphantom{*} & 0.35 ± 0.01\hphantom{*}\hphantom{*} & 2.59 ± 0.12\hphantom{*}\hphantom{*} & 0.09 ± 0.01\hphantom{*}\hphantom{*} & 0.99 ± 0.01*\hphantom{*}\\

 & ECCCo (no CP) & 1.86 ± 0.12\hphantom{*}\hphantom{*} & 1.00 ± 0.18\hphantom{*}\hphantom{*} & 0.36 ± 0.01\hphantom{*}\hphantom{*} & 1.06 ± 0.05\hphantom{*}\hphantom{*} & 0.13 ± 0.01\hphantom{*}\hphantom{*} & \textbf{0.99 ± 0.01}*\hphantom{*}\\

 & ECCCo (no EBM) & 1.91 ± 0.11\hphantom{*}\hphantom{*} & 1.09 ± 0.18\hphantom{*}\hphantom{*} & 0.31 ± 0.01*\hphantom{*} & 0.89 ± 0.03\hphantom{*}\hphantom{*} & 0.14 ± 0.01\hphantom{*}\hphantom{*} & 0.98 ± 0.01\hphantom{*}\hphantom{*}\\
\bottomrule
\end{tabular}}
\end{table}

%% file: contents/table-german-credit.tex
\begin{table}

\caption{All results for German Credit dataset: sample averages +/- one standard deviation over all counterfactuals. Best outcomes are highlighted in bold. Asterisks indicate that the given value is more than one (*) or two (**) standard deviations away from the baseline (\textit{Wachter}). \label{tab:results-german-credit} \newline}
\centering
\resizebox{\linewidth}{!}{
\begin{tabular}[t]{cccccccc}
\toprule
Model & Generator & Unfaithfulness ↓ & Implausibility ↓ & Uncertainty ↓ & Cost ↓ & Redundancy ↑ & Validity ↑\\
\midrule
 & ECCCo-L1 & 1.69 ± 0.03** & 5.16 ± 0.08\hphantom{*}\hphantom{*} & 0.59 ± 0.01** & 3.06 ± 0.19\hphantom{*}\hphantom{*} & 0.29 ± 0.01\hphantom{*}\hphantom{*} & 1.00 ± 0.01\hphantom{*}\hphantom{*}\\

 & ECCCo-L1 (no CP) & \textbf{1.69 ± 0.03}** & 5.16 ± 0.07\hphantom{*}\hphantom{*} & 0.60 ± 0.01\hphantom{*}\hphantom{*} & 2.98 ± 0.19\hphantom{*}\hphantom{*} & 0.29 ± 0.01\hphantom{*}\hphantom{*} & 1.00 ± 0.01\hphantom{*}\hphantom{*}\\

 & ECCCo-L1 (no EBM) & 1.70 ± 0.04** & 5.18 ± 0.08\hphantom{*}\hphantom{*} & 0.59 ± 0.01** & 3.04 ± 0.19\hphantom{*}\hphantom{*} & 0.30 ± 0.01\hphantom{*}\hphantom{*} & \textbf{1.00 ± 0.01}\hphantom{*}\hphantom{*}\\

 & ECCCo & 4.47 ± 0.09*\hphantom{*} & 5.07 ± 0.07*\hphantom{*} & 0.60 ± 0.01\hphantom{*}\hphantom{*} & 3.65 ± 0.28\hphantom{*}\hphantom{*} & 0.27 ± 0.01\hphantom{*}\hphantom{*} & 1.00 ± 0.01\hphantom{*}\hphantom{*}\\

 & ECCCo+ & 4.08 ± 0.07** & \textbf{3.82 ± 0.06}** & 0.22 ± 0.01** & 13.04 ± 0.29\hphantom{*}\hphantom{*} & 0.09 ± 0.01\hphantom{*}\hphantom{*} & 0.39 ± 0.01\hphantom{*}\hphantom{*}\\

 & ECCCo (no CP) & 4.48 ± 0.10\hphantom{*}\hphantom{*} & 5.06 ± 0.07*\hphantom{*} & 0.61 ± 0.01\hphantom{*}\hphantom{*} & 3.59 ± 0.28\hphantom{*}\hphantom{*} & 0.27 ± 0.01\hphantom{*}\hphantom{*} & 1.00 ± 0.01\hphantom{*}\hphantom{*}\\

 & ECCCo (no EBM) & 4.55 ± 0.09\hphantom{*}\hphantom{*} & 5.18 ± 0.08\hphantom{*}\hphantom{*} & 0.59 ± 0.01** & 3.04 ± 0.19\hphantom{*}\hphantom{*} & 0.30 ± 0.01\hphantom{*}\hphantom{*} & \textbf{1.00 ± 0.01}\hphantom{*}\hphantom{*}\\
\bottomrule
\end{tabular}}
\end{table}

%% file: contents/table-mnist.tex
\begin{table}

\caption{All results for MNIST dataset: sample averages +/- one standard deviation over all counterfactuals. Best outcomes are highlighted in bold. Asterisks indicate that the given value is more than one (*) or two (**) standard deviations away from the baseline (\textit{Wachter}). \label{tab:results-mnist} \newline}
\centering
\resizebox{\linewidth}{!}{
\begin{tabular}[t]{cccccccc}
\toprule
Model & Generator & Unfaithfulness ↓ & Implausibility ↓ & Uncertainty ↓ & Cost ↓ & Redundancy ↑ & Validity ↑\\
\midrule
 & ECCCo-L1 & \textbf{0.10 ± 0.00}** & 0.35 ± 0.00\hphantom{*}\hphantom{*} & 4.59 ± 0.01** & 23.75 ± 0.97\hphantom{*}\hphantom{*} & 0.00 ± 0.00\hphantom{*}\hphantom{*} & \textbf{1.00 ± 0.00}\hphantom{*}\hphantom{*}\\

 & ECCCo & 0.25 ± 0.00** & 0.46 ± 0.00\hphantom{*}\hphantom{*} & 4.50 ± 0.00** & 160.43 ± 1.31\hphantom{*}\hphantom{*} & 0.00 ± 0.00\hphantom{*}\hphantom{*} & \textbf{1.00 ± 0.00}\hphantom{*}\hphantom{*}\\

 & ECCCo+ & 0.25 ± 0.00** & 0.31 ± 0.00** & 4.51 ± 0.00** & 196.62 ± 3.62\hphantom{*}\hphantom{*} & 0.00 ± 0.00\hphantom{*}\hphantom{*} & \textbf{1.00 ± 0.00}\hphantom{*}\hphantom{*}\\

 & REVISE & 0.25 ± 0.00\hphantom{*}\hphantom{*} & \textbf{0.29 ± 0.00}** & 3.82 ± 0.14** & 239.71 ± 6.72\hphantom{*}\hphantom{*} & 0.00 ± 0.00\hphantom{*}\hphantom{*} & 0.96 ± 0.02\hphantom{*}\hphantom{*}\\

 & Schut & 0.25 ± 0.00\hphantom{*}\hphantom{*} & 0.34 ± 0.00** & \textbf{0.73 ± 0.15}** & \textbf{7.01 ± 0.18}** & \textbf{0.99 ± 0.00}** & 0.16 ± 0.03\hphantom{*}\hphantom{*}\\

\multirow[t]{-6}{*}{\centering\arraybackslash JEM} & Wachter & 0.25 ± 0.00\hphantom{*}\hphantom{*} & 0.35 ± 0.00\hphantom{*}\hphantom{*} & 4.62 ± 0.01\hphantom{*}\hphantom{*} & 22.90 ± 0.99\hphantom{*}\hphantom{*} & 0.00 ± 0.00\hphantom{*}\hphantom{*} & \textbf{1.00 ± 0.00}\hphantom{*}\hphantom{*}\\
\cmidrule{1-8}
 & ECCCo-L1 & \textbf{0.10 ± 0.00}** & 0.34 ± 0.00\hphantom{*}\hphantom{*} & 2.44 ± 0.01** & 35.89 ± 1.47\hphantom{*}\hphantom{*} & 0.00 ± 0.00\hphantom{*}\hphantom{*} & \textbf{1.00 ± 0.00}\hphantom{*}\hphantom{*}\\
\bottomrule
\end{tabular}}
\end{table}

%% file: contents/table-fashion-mnist.tex
\begin{table}

\caption{All results for Fashion MNIST dataset: sample averages +/- one standard deviation over all counterfactuals. Best outcomes are highlighted in bold. Asterisks indicate that the given value is more than one (*) or two (**) standard deviations away from the baseline (\textit{Wachter}). \label{tab:results-fashion-mnist} \newline}
\centering
\resizebox{\linewidth}{!}{
\begin{tabular}[t]{cccccccc}
\toprule
Model & Generator & Unfaithfulness ↓ & Implausibility ↓ & Uncertainty ↓ & Cost ↓ & Redundancy ↑ & Validity ↑\\
\midrule
 & ECCCo-L1 & \textbf{0.10 ± 0.00}** & 0.40 ± 0.00\hphantom{*}\hphantom{*} & 4.15 ± 0.10\hphantom{*}\hphantom{*} & 31.58 ± 1.34\hphantom{*}\hphantom{*} & 0.00 ± 0.00\hphantom{*}\hphantom{*} & 0.84 ± 0.02*\hphantom{*}\\

 & ECCCo & 0.25 ± 0.00** & 0.43 ± 0.00\hphantom{*}\hphantom{*} & 4.56 ± 0.02\hphantom{*}\hphantom{*} & 79.06 ± 1.35\hphantom{*}\hphantom{*} & 0.00 ± 0.00** & \textbf{1.00 ± 0.00}**\\

 & ECCCo+ & 0.25 ± 0.00** & 0.38 ± 0.00** & 2.98 ± 0.15** & 184.42 ± 4.53\hphantom{*}\hphantom{*} & 0.00 ± 0.00\hphantom{*}\hphantom{*} & 0.62 ± 0.03\hphantom{*}\hphantom{*}\\

 & REVISE & 0.25 ± 0.00\hphantom{*}\hphantom{*} & \textbf{0.37 ± 0.00}** & \textbf{0.74 ± 0.17}** & 196.14 ± 4.92\hphantom{*}\hphantom{*} & 0.00 ± 0.00\hphantom{*}\hphantom{*} & 0.16 ± 0.04\hphantom{*}\hphantom{*}\\

 & Schut & 0.25 ± 0.00\hphantom{*}\hphantom{*} & 0.40 ± 0.00*\hphantom{*} & 0.88 ± 0.13** & \textbf{8.21 ± 0.10}** & \textbf{0.99 ± 0.00}** & 0.18 ± 0.03\hphantom{*}\hphantom{*}\\

\multirow[t]{-6}{*}{\centering\arraybackslash JEM} & Wachter & 0.25 ± 0.00\hphantom{*}\hphantom{*} & 0.40 ± 0.00\hphantom{*}\hphantom{*} & 4.07 ± 0.11\hphantom{*}\hphantom{*} & 29.91 ± 1.30\hphantom{*}\hphantom{*} & 0.00 ± 0.00\hphantom{*}\hphantom{*} & 0.81 ± 0.02\hphantom{*}\hphantom{*}\\
\cmidrule{1-8}
 & ECCCo-L1 & \textbf{0.10 ± 0.00}** & 0.40 ± 0.00\hphantom{*}\hphantom{*} & 0.79 ± 0.10\hphantom{*}\hphantom{*} & 40.74 ± 1.07\hphantom{*}\hphantom{*} & 0.00 ± 0.00\hphantom{*}\hphantom{*} & 0.26 ± 0.04*\hphantom{*}\\
\bottomrule
\end{tabular}}
\end{table}

%% file: contents/table-linearly-separable-valid.tex
\begin{table}

\caption{All results for Linearly Separable dataset: sample averages +/- one standard deviation over all valid counterfactuals. Best outcomes are highlighted in bold. Asterisks indicate that the given value is more than one (*) or two (**) standard deviations away from the baseline (\textit{Wachter}). \label{tab:results-linearly-separable-valid} \newline}
\centering
\resizebox{\linewidth}{!}{
\begin{tabular}[t]{cccccccc}
\toprule
Model & Generator & Unfaithfulness ↓ & Implausibility ↓ & Uncertainty ↓ & Cost ↓ & Redundancy ↑ & Validity ↑\\
\midrule
 & ECCCo-L1 & 0.06 ± 0.01** & \textbf{0.16 ± 0.02}** & 0.00 ± 0.00\hphantom{*}\hphantom{*} & 0.93 ± 0.02\hphantom{*}\hphantom{*} & 0.00 ± 0.00\hphantom{*}\hphantom{*} & 1.00 ± 0.00\hphantom{*}\hphantom{*}\\

 & ECCCo-L1 (no CP) & \textbf{0.06 ± 0.01}** & 0.16 ± 0.02** & 0.00 ± 0.00\hphantom{*}\hphantom{*} & 0.93 ± 0.02\hphantom{*}\hphantom{*} & 0.00 ± 0.00\hphantom{*}\hphantom{*} & 1.00 ± 0.00\hphantom{*}\hphantom{*}\\

 & ECCCo-L1 (no EBM) & 0.09 ± 0.01** & 0.24 ± 0.01\hphantom{*}\hphantom{*} & 0.00 ± 0.00\hphantom{*}\hphantom{*} & 0.88 ± 0.01\hphantom{*}\hphantom{*} & 0.00 ± 0.00\hphantom{*}\hphantom{*} & 1.00 ± 0.00\hphantom{*}\hphantom{*}\\

 & ECCCo & 0.19 ± 0.01** & 0.19 ± 0.01** & 0.00 ± 0.00\hphantom{*}\hphantom{*} & 0.85 ± 0.01** & \textbf{0.00 ± 0.00}\hphantom{*}\hphantom{*} & 1.00 ± 0.00\hphantom{*}\hphantom{*}\\

 & ECCCo (no CP) & 0.19 ± 0.01** & 0.19 ± 0.01** & 0.00 ± 0.00\hphantom{*}\hphantom{*} & 0.85 ± 0.01** & \textbf{0.00 ± 0.00}\hphantom{*}\hphantom{*} & 1.00 ± 0.00\hphantom{*}\hphantom{*}\\

 & ECCCo (no EBM) & 0.24 ± 0.01\hphantom{*}\hphantom{*} & 0.24 ± 0.01\hphantom{*}\hphantom{*} & 0.00 ± 0.00\hphantom{*}\hphantom{*} & 0.88 ± 0.01\hphantom{*}\hphantom{*} & 0.00 ± 0.00\hphantom{*}\hphantom{*} & 1.00 ± 0.00\hphantom{*}\hphantom{*}\\

 & REVISE & 0.24 ± 0.02\hphantom{*}\hphantom{*} & 0.17 ± 0.00** & 0.00 ± 0.00\hphantom{*}\hphantom{*} & \textbf{0.42 ± 0.02}** & 0.00 ± 0.00\hphantom{*}\hphantom{*} & 1.00 ± 0.00\hphantom{*}\hphantom{*}\\
\bottomrule
\end{tabular}}
\end{table}

%% file: contents/table-circles-valid.tex
\begin{table}

\caption{All results for Circles dataset: sample averages +/- one standard deviation over all valid counterfactuals. Best outcomes are highlighted in bold. Asterisks indicate that the given value is more than one (*) or two (**) standard deviations away from the baseline (\textit{Wachter}). \label{tab:results-circles-valid} \newline}
\centering
\resizebox{\linewidth}{!}{
\begin{tabular}[t]{cccccccc}
\toprule
Model & Generator & Unfaithfulness ↓ & Implausibility ↓ & Uncertainty ↓ & Cost ↓ & Redundancy ↑ & Validity ↑\\
\midrule
 & ECCCo-L1 & 0.24 ± 0.01** & 0.33 ± 0.01\hphantom{*}\hphantom{*} & 0.00 ± 0.00\hphantom{*}\hphantom{*} & 0.50 ± 0.02\hphantom{*}\hphantom{*} & 0.00 ± 0.00\hphantom{*}\hphantom{*} & 1.00 ± 0.00\hphantom{*}\hphantom{*}\\

 & ECCCo-L1 (no CP) & 0.24 ± 0.02** & 0.33 ± 0.01\hphantom{*}\hphantom{*} & 0.00 ± 0.00\hphantom{*}\hphantom{*} & 0.50 ± 0.02\hphantom{*}\hphantom{*} & 0.00 ± 0.00\hphantom{*}\hphantom{*} & 1.00 ± 0.00\hphantom{*}\hphantom{*}\\

 & ECCCo-L1 (no EBM) & 0.25 ± 0.02** & 0.33 ± 0.01\hphantom{*}\hphantom{*} & 0.00 ± 0.00\hphantom{*}\hphantom{*} & 0.50 ± 0.02\hphantom{*}\hphantom{*} & 0.00 ± 0.00\hphantom{*}\hphantom{*} & 1.00 ± 0.00\hphantom{*}\hphantom{*}\\

 & ECCCo & 0.61 ± 0.05\hphantom{*}\hphantom{*} & 0.33 ± 0.01\hphantom{*}\hphantom{*} & 0.00 ± 0.00\hphantom{*}\hphantom{*} & 0.62 ± 0.02\hphantom{*}\hphantom{*} & 0.00 ± 0.00\hphantom{*}\hphantom{*} & 1.00 ± 0.00\hphantom{*}\hphantom{*}\\

 & ECCCo (no CP) & 0.60 ± 0.05\hphantom{*}\hphantom{*} & 0.33 ± 0.01\hphantom{*}\hphantom{*} & 0.00 ± 0.00\hphantom{*}\hphantom{*} & 0.62 ± 0.02\hphantom{*}\hphantom{*} & 0.00 ± 0.00\hphantom{*}\hphantom{*} & 1.00 ± 0.00\hphantom{*}\hphantom{*}\\

 & ECCCo (no EBM) & 0.62 ± 0.04\hphantom{*}\hphantom{*} & 0.33 ± 0.01\hphantom{*}\hphantom{*} & 0.00 ± 0.00\hphantom{*}\hphantom{*} & 0.50 ± 0.02\hphantom{*}\hphantom{*} & 0.00 ± 0.00\hphantom{*}\hphantom{*} & 1.00 ± 0.00\hphantom{*}\hphantom{*}\\

 & REVISE & \textbf{0.12 ± 0.02}** & 0.43 ± 0.00\hphantom{*}\hphantom{*} & 0.00 ± 0.00\hphantom{*}\hphantom{*} & 1.27 ± 0.02\hphantom{*}\hphantom{*} & 0.00 ± 0.00\hphantom{*}\hphantom{*} & 1.00 ± 0.00\hphantom{*}\hphantom{*}\\
\bottomrule
\end{tabular}}
\end{table}

%% file: contents/table-moons-valid.tex
\begin{table}

\caption{All results for Moons dataset: sample averages +/- one standard deviation over all valid counterfactuals. Best outcomes are highlighted in bold. Asterisks indicate that the given value is more than one (*) or two (**) standard deviations away from the baseline (\textit{Wachter}). \label{tab:results-moons-valid} \newline}
\centering
\resizebox{\linewidth}{!}{
\begin{tabular}[t]{cccccccc}
\toprule
Model & Generator & Unfaithfulness ↓ & Implausibility ↓ & Uncertainty ↓ & Cost ↓ & Redundancy ↑ & Validity ↑\\
\midrule
 & ECCCo-L1 & 0.29 ± 0.02** & 0.42 ± 0.04\hphantom{*}\hphantom{*} & 0.14 ± 0.03\hphantom{*}\hphantom{*} & 1.02 ± 0.03\hphantom{*}\hphantom{*} & 0.00 ± 0.00\hphantom{*}\hphantom{*} & 1.00 ± 0.00\hphantom{*}\hphantom{*}\\

 & ECCCo-L1 (no CP) & \textbf{0.29 ± 0.02}** & 0.42 ± 0.04\hphantom{*}\hphantom{*} & 0.15 ± 0.03\hphantom{*}\hphantom{*} & \textbf{1.02 ± 0.03}\hphantom{*}\hphantom{*} & 0.00 ± 0.00\hphantom{*}\hphantom{*} & 1.00 ± 0.00\hphantom{*}\hphantom{*}\\

 & ECCCo-L1 (no EBM) & 0.30 ± 0.02** & 0.42 ± 0.04\hphantom{*}\hphantom{*} & 0.13 ± 0.03\hphantom{*}\hphantom{*} & 1.03 ± 0.03\hphantom{*}\hphantom{*} & 0.00 ± 0.00\hphantom{*}\hphantom{*} & 1.00 ± 0.00\hphantom{*}\hphantom{*}\\

 & ECCCo & 0.55 ± 0.06** & 0.15 ± 0.01** & \textbf{0.01 ± 0.01}** & 1.28 ± 0.04\hphantom{*}\hphantom{*} & 0.00 ± 0.00\hphantom{*}\hphantom{*} & 1.00 ± 0.00\hphantom{*}\hphantom{*}\\

 & ECCCo (no CP) & 0.57 ± 0.05** & \textbf{0.15 ± 0.01}** & 0.01 ± 0.01** & 1.28 ± 0.04\hphantom{*}\hphantom{*} & 0.00 ± 0.00\hphantom{*}\hphantom{*} & 1.00 ± 0.00\hphantom{*}\hphantom{*}\\

 & ECCCo (no EBM) & 0.69 ± 0.04\hphantom{*}\hphantom{*} & 0.42 ± 0.04\hphantom{*}\hphantom{*} & 0.13 ± 0.03\hphantom{*}\hphantom{*} & 1.03 ± 0.03\hphantom{*}\hphantom{*} & 0.00 ± 0.00\hphantom{*}\hphantom{*} & 1.00 ± 0.00\hphantom{*}\hphantom{*}\\

 & REVISE & 0.54 ± 0.05** & 0.27 ± 0.01** & 0.14 ± 0.05\hphantom{*}\hphantom{*} & 1.42 ± 0.13\hphantom{*}\hphantom{*} & 0.00 ± 0.00\hphantom{*}\hphantom{*} & 1.00 ± 0.00\hphantom{*}\hphantom{*}\\
\bottomrule
\end{tabular}}
\end{table}

%% file: contents/table-california-housing-valid.tex
\begin{table}

\caption{All results for California Housing dataset: sample averages +/- one standard deviation over all valid counterfactuals. Best outcomes are highlighted in bold. Asterisks indicate that the given value is more than one (*) or two (**) standard deviations away from the baseline (\textit{Wachter}). \label{tab:results-california-housing-valid} \newline}
\centering
\resizebox{\linewidth}{!}{
\begin{tabular}[t]{cccccccc}
\toprule
Model & Generator & Unfaithfulness ↓ & Implausibility ↓ & Uncertainty ↓ & Cost ↓ & Redundancy ↑ & Validity ↑\\
\midrule
 & ECCCo-L1 & 0.98 ± 0.05** & 0.99 ± 0.09\hphantom{*}\hphantom{*} & 0.25 ± 0.02\hphantom{*}\hphantom{*} & 1.20 ± 0.07\hphantom{*}\hphantom{*} & 0.00 ± 0.00\hphantom{*}\hphantom{*} & 1.00 ± 0.00\hphantom{*}\hphantom{*}\\

 & ECCCo-L1 (no CP) & \textbf{0.97 ± 0.05}** & 0.99 ± 0.09\hphantom{*}\hphantom{*} & 0.26 ± 0.02\hphantom{*}\hphantom{*} & 1.15 ± 0.06\hphantom{*}\hphantom{*} & 0.00 ± 0.00\hphantom{*}\hphantom{*} & 1.00 ± 0.00\hphantom{*}\hphantom{*}\\

 & ECCCo-L1 (no EBM) & 1.00 ± 0.05** & 1.02 ± 0.09\hphantom{*}\hphantom{*} & 0.25 ± 0.01\hphantom{*}\hphantom{*} & 1.10 ± 0.07\hphantom{*}\hphantom{*} & 0.01 ± 0.00\hphantom{*}\hphantom{*} & 1.00 ± 0.00\hphantom{*}\hphantom{*}\\

 & ECCCo & 1.93 ± 0.13*\hphantom{*} & 0.77 ± 0.07** & \textbf{0.18 ± 0.00}** & 2.05 ± 0.16\hphantom{*}\hphantom{*} & 0.01 ± 0.00\hphantom{*}\hphantom{*} & 1.00 ± 0.00\hphantom{*}\hphantom{*}\\

 & ECCCo+ & 1.83 ± 0.12** & 0.69 ± 0.04** & 0.19 ± 0.00** & 3.27 ± 0.18\hphantom{*}\hphantom{*} & 0.00 ± 0.00\hphantom{*}\hphantom{*} & 1.00 ± 0.00\hphantom{*}\hphantom{*}\\

 & ECCCo (no CP) & 1.95 ± 0.15*\hphantom{*} & 0.77 ± 0.07** & 0.18 ± 0.00** & 2.03 ± 0.16\hphantom{*}\hphantom{*} & 0.00 ± 0.00\hphantom{*}\hphantom{*} & 1.00 ± 0.00\hphantom{*}\hphantom{*}\\

 & ECCCo (no EBM) & 2.11 ± 0.11\hphantom{*}\hphantom{*} & 1.02 ± 0.09\hphantom{*}\hphantom{*} & 0.25 ± 0.01\hphantom{*}\hphantom{*} & 1.10 ± 0.07\hphantom{*}\hphantom{*} & 0.01 ± 0.00\hphantom{*}\hphantom{*} & 1.00 ± 0.00\hphantom{*}\hphantom{*}\\
\bottomrule
\end{tabular}}
\end{table}

%% file: contents/table-gmsc-valid.tex
\begin{table}

\caption{All results for GMSC dataset: sample averages +/- one standard deviation over all valid counterfactuals. Best outcomes are highlighted in bold. Asterisks indicate that the given value is more than one (*) or two (**) standard deviations away from the baseline (\textit{Wachter}). \label{tab:results-gmsc-valid} \newline}
\centering
\resizebox{\linewidth}{!}{
\begin{tabular}[t]{cccccccc}
\toprule
Model & Generator & Unfaithfulness ↓ & Implausibility ↓ & Uncertainty ↓ & Cost ↓ & Redundancy ↑ & Validity ↑\\
\midrule
 & ECCCo-L1 & \textbf{0.92 ± 0.04}** & 0.94 ± 0.09\hphantom{*}\hphantom{*} & 0.30 ± 0.01** & 1.02 ± 0.04\hphantom{*}\hphantom{*} & 0.13 ± 0.01\hphantom{*}\hphantom{*} & 1.00 ± 0.00\hphantom{*}\hphantom{*}\\

 & ECCCo-L1 (no CP) & 0.93 ± 0.04** & 0.93 ± 0.09\hphantom{*}\hphantom{*} & 0.31 ± 0.01*\hphantom{*} & 0.94 ± 0.04\hphantom{*}\hphantom{*} & 0.14 ± 0.01\hphantom{*}\hphantom{*} & 1.00 ± 0.00\hphantom{*}\hphantom{*}\\

 & ECCCo-L1 (no EBM) & 0.95 ± 0.04** & 0.93 ± 0.10\hphantom{*}\hphantom{*} & 0.31 ± 0.01*\hphantom{*} & 0.88 ± 0.03\hphantom{*}\hphantom{*} & 0.14 ± 0.01\hphantom{*}\hphantom{*} & 1.00 ± 0.00\hphantom{*}\hphantom{*}\\

 & ECCCo & 1.78 ± 0.10\hphantom{*}\hphantom{*} & 0.86 ± 0.09\hphantom{*}\hphantom{*} & 0.35 ± 0.01\hphantom{*}\hphantom{*} & 1.13 ± 0.05\hphantom{*}\hphantom{*} & 0.13 ± 0.01\hphantom{*}\hphantom{*} & 1.00 ± 0.00\hphantom{*}\hphantom{*}\\

 & ECCCo+ & 1.65 ± 0.10** & 0.73 ± 0.05** & 0.35 ± 0.01\hphantom{*}\hphantom{*} & 2.58 ± 0.11\hphantom{*}\hphantom{*} & 0.09 ± 0.01\hphantom{*}\hphantom{*} & 1.00 ± 0.00\hphantom{*}\hphantom{*}\\

 & ECCCo (no CP) & 1.79 ± 0.10\hphantom{*}\hphantom{*} & 0.85 ± 0.09\hphantom{*}\hphantom{*} & 0.36 ± 0.01\hphantom{*}\hphantom{*} & 1.06 ± 0.05\hphantom{*}\hphantom{*} & 0.13 ± 0.01\hphantom{*}\hphantom{*} & 1.00 ± 0.00\hphantom{*}\hphantom{*}\\

 & ECCCo (no EBM) & 1.84 ± 0.10\hphantom{*}\hphantom{*} & 0.93 ± 0.10\hphantom{*}\hphantom{*} & 0.31 ± 0.01*\hphantom{*} & 0.88 ± 0.03\hphantom{*}\hphantom{*} & 0.14 ± 0.01\hphantom{*}\hphantom{*} & 1.00 ± 0.00\hphantom{*}\hphantom{*}\\
\bottomrule
\end{tabular}}
\end{table}

%% file: contents/table-german-credit-valid.tex
\begin{table}

\caption{All results for German Credit dataset: sample averages +/- one standard deviation over all valid counterfactuals. Best outcomes are highlighted in bold. Asterisks indicate that the given value is more than one (*) or two (**) standard deviations away from the baseline (\textit{Wachter}). \label{tab:results-german-credit-valid} \newline}
\centering
\resizebox{\linewidth}{!}{
\begin{tabular}[t]{cccccccc}
\toprule
Model & Generator & Unfaithfulness ↓ & Implausibility ↓ & Uncertainty ↓ & Cost ↓ & Redundancy ↑ & Validity ↑\\
\midrule
 & ECCCo-L1 & 1.69 ± 0.03** & 5.16 ± 0.07\hphantom{*}\hphantom{*} & 0.59 ± 0.00** & 3.04 ± 0.20\hphantom{*}\hphantom{*} & 0.29 ± 0.01\hphantom{*}\hphantom{*} & 1.00 ± 0.00\hphantom{*}\hphantom{*}\\

 & ECCCo-L1 (no CP) & \textbf{1.69 ± 0.03}** & 5.16 ± 0.07\hphantom{*}\hphantom{*} & 0.60 ± 0.00\hphantom{*}\hphantom{*} & 2.96 ± 0.19\hphantom{*}\hphantom{*} & 0.29 ± 0.01\hphantom{*}\hphantom{*} & 1.00 ± 0.00\hphantom{*}\hphantom{*}\\

 & ECCCo-L1 (no EBM) & 1.70 ± 0.04** & 5.18 ± 0.08\hphantom{*}\hphantom{*} & 0.59 ± 0.00** & 3.02 ± 0.19\hphantom{*}\hphantom{*} & 0.30 ± 0.01\hphantom{*}\hphantom{*} & 1.00 ± 0.00\hphantom{*}\hphantom{*}\\

 & ECCCo & 4.48 ± 0.09*\hphantom{*} & 5.07 ± 0.07*\hphantom{*} & 0.60 ± 0.01\hphantom{*}\hphantom{*} & 3.63 ± 0.28\hphantom{*}\hphantom{*} & 0.27 ± 0.01\hphantom{*}\hphantom{*} & 1.00 ± 0.00\hphantom{*}\hphantom{*}\\

 & ECCCo+ & 4.85 ± 0.10\hphantom{*}\hphantom{*} & 3.76 ± 0.10** & 0.55 ± 0.01** & 12.77 ± 0.40\hphantom{*}\hphantom{*} & 0.08 ± 0.01\hphantom{*}\hphantom{*} & 1.00 ± 0.00\hphantom{*}\hphantom{*}\\

 & ECCCo (no CP) & 4.48 ± 0.10\hphantom{*}\hphantom{*} & 5.07 ± 0.07*\hphantom{*} & 0.62 ± 0.01\hphantom{*}\hphantom{*} & 3.58 ± 0.28\hphantom{*}\hphantom{*} & 0.27 ± 0.01\hphantom{*}\hphantom{*} & 1.00 ± 0.00\hphantom{*}\hphantom{*}\\

 & ECCCo (no EBM) & 4.55 ± 0.09\hphantom{*}\hphantom{*} & 5.18 ± 0.08\hphantom{*}\hphantom{*} & 0.59 ± 0.00** & 3.02 ± 0.19\hphantom{*}\hphantom{*} & 0.30 ± 0.01\hphantom{*}\hphantom{*} & 1.00 ± 0.00\hphantom{*}\hphantom{*}\\
\bottomrule
\end{tabular}}
\end{table}

%% file: contents/table-mnist-valid.tex
\begin{table}

\caption{All results for MNIST dataset: sample averages +/- one standard deviation over all valid counterfactuals. Best outcomes are highlighted in bold. Asterisks indicate that the given value is more than one (*) or two (**) standard deviations away from the baseline (\textit{Wachter}). \label{tab:results-mnist-valid} \newline}
\centering
\resizebox{\linewidth}{!}{
\begin{tabular}[t]{cccccccc}
\toprule
Model & Generator & Unfaithfulness ↓ & Implausibility ↓ & Uncertainty ↓ & Cost ↓ & Redundancy ↑ & Validity ↑\\
\midrule
 & ECCCo-L1 & \textbf{0.10 ± 0.00}** & 0.35 ± 0.00\hphantom{*}\hphantom{*} & 4.59 ± 0.01** & 23.75 ± 0.97\hphantom{*}\hphantom{*} & 0.00 ± 0.00\hphantom{*}\hphantom{*} & 1.00 ± 0.00\hphantom{*}\hphantom{*}\\

 & ECCCo & 0.25 ± 0.00** & 0.46 ± 0.00\hphantom{*}\hphantom{*} & 4.50 ± 0.00** & 160.43 ± 1.31\hphantom{*}\hphantom{*} & 0.00 ± 0.00\hphantom{*}\hphantom{*} & 1.00 ± 0.00\hphantom{*}\hphantom{*}\\

 & ECCCo+ & 0.25 ± 0.00** & 0.31 ± 0.00** & 4.51 ± 0.00** & 196.62 ± 3.62\hphantom{*}\hphantom{*} & 0.00 ± 0.00\hphantom{*}\hphantom{*} & 1.00 ± 0.00\hphantom{*}\hphantom{*}\\

 & REVISE & 0.25 ± 0.00\hphantom{*}\hphantom{*} & \textbf{0.29 ± 0.00}** & \textbf{3.90 ± 0.14}** & 240.16 ± 6.54\hphantom{*}\hphantom{*} & 0.00 ± 0.00\hphantom{*}\hphantom{*} & 1.00 ± 0.00\hphantom{*}\hphantom{*}\\

 & Schut & 0.25 ± 0.00\hphantom{*}\hphantom{*} & 0.32 ± 0.01** & 4.61 ± 0.05\hphantom{*}\hphantom{*} & \textbf{6.02 ± 0.60}** & \textbf{0.99 ± 0.00}** & 1.00 ± 0.00\hphantom{*}\hphantom{*}\\

\multirow[t]{-6}{*}{\centering\arraybackslash JEM} & Wachter & 0.25 ± 0.00\hphantom{*}\hphantom{*} & 0.35 ± 0.00\hphantom{*}\hphantom{*} & 4.62 ± 0.01\hphantom{*}\hphantom{*} & 22.90 ± 0.99\hphantom{*}\hphantom{*} & 0.00 ± 0.00\hphantom{*}\hphantom{*} & 1.00 ± 0.00\hphantom{*}\hphantom{*}\\
\cmidrule{1-8}
 & ECCCo-L1 & \textbf{0.10 ± 0.00}** & 0.34 ± 0.00\hphantom{*}\hphantom{*} & 2.44 ± 0.01** & 35.89 ± 1.47\hphantom{*}\hphantom{*} & 0.00 ± 0.00\hphantom{*}\hphantom{*} & 1.00 ± 0.00\hphantom{*}\hphantom{*}\\
\bottomrule
\end{tabular}}
\end{table}

%% file: contents/table-fashion-mnist-valid.tex
\begin{table}

\caption{All results for Fashion MNIST dataset: sample averages +/- one standard deviation over all valid counterfactuals. Best outcomes are highlighted in bold. Asterisks indicate that the given value is more than one (*) or two (**) standard deviations away from the baseline (\textit{Wachter}). \label{tab:results-fashion-mnist-valid} \newline}
\centering
\resizebox{\linewidth}{!}{
\begin{tabular}[t]{cccccccc}
\toprule
Model & Generator & Unfaithfulness ↓ & Implausibility ↓ & Uncertainty ↓ & Cost ↓ & Redundancy ↑ & Validity ↑\\
\midrule
 & ECCCo-L1 & \textbf{0.10 ± 0.00}** & 0.39 ± 0.01\hphantom{*}\hphantom{*} & 4.95 ± 0.01** & 29.93 ± 1.60\hphantom{*}\hphantom{*} & 0.00 ± 0.00\hphantom{*}\hphantom{*} & 1.00 ± 0.00\hphantom{*}\hphantom{*}\\

 & ECCCo & 0.25 ± 0.00** & 0.43 ± 0.00\hphantom{*}\hphantom{*} & 4.57 ± 0.01** & 78.99 ± 1.28\hphantom{*}\hphantom{*} & 0.00 ± 0.00** & 1.00 ± 0.00\hphantom{*}\hphantom{*}\\

 & ECCCo+ & 0.25 ± 0.00** & 0.37 ± 0.00** & 4.79 ± 0.04** & 175.11 ± 4.20\hphantom{*}\hphantom{*} & 0.00 ± 0.00\hphantom{*}\hphantom{*} & 1.00 ± 0.00\hphantom{*}\hphantom{*}\\

 & REVISE & 0.25 ± 0.00*\hphantom{*} & \textbf{0.30 ± 0.02}** & \textbf{3.82 ± 0.49}** & 208.22 ± 16.18\hphantom{*}\hphantom{*} & 0.00 ± 0.00\hphantom{*}\hphantom{*} & 1.00 ± 0.00\hphantom{*}\hphantom{*}\\

 & Schut & 0.25 ± 0.00\hphantom{*}\hphantom{*} & 0.34 ± 0.01** & 4.78 ± 0.04** & \textbf{8.29 ± 0.23}** & \textbf{0.99 ± 0.00}** & 1.00 ± 0.00\hphantom{*}\hphantom{*}\\

\multirow[t]{-6}{*}{\centering\arraybackslash JEM} & Wachter & 0.25 ± 0.00\hphantom{*}\hphantom{*} & 0.39 ± 0.01\hphantom{*}\hphantom{*} & 5.01 ± 0.02\hphantom{*}\hphantom{*} & 28.33 ± 1.56\hphantom{*}\hphantom{*} & 0.00 ± 0.00\hphantom{*}\hphantom{*} & 1.00 ± 0.00\hphantom{*}\hphantom{*}\\
\cmidrule{1-8}
 & ECCCo-L1 & \textbf{0.10 ± 0.00}** & 0.37 ± 0.01\hphantom{*}\hphantom{*} & 2.98 ± 0.04\hphantom{*}\hphantom{*} & 37.85 ± 3.04\hphantom{*}\hphantom{*} & 0.00 ± 0.00\hphantom{*}\hphantom{*} & 1.00 ± 0.00\hphantom{*}\hphantom{*}\\
\bottomrule
\end{tabular}}
\end{table}